\documentclass[oneside,phd,etd]{WSUclass}
\usepackage{import}
\usepackage{preamble}
\usepackage[utf8]{inputenc} 
\usepackage[T1]{fontenc} 
\usepackage{hyperref}
\usepackage{comment}
\usepackage{framed}
\usepackage{booktabs}
\usepackage{graphicx}
\usepackage{amssymb}
\usepackage{amstext}
\usepackage{tabularx}
\usepackage{amsmath}
\usepackage[square,numbers]{natbib}
\usepackage{textcomp}
\usepackage{ bbold }
\usepackage{xpatch}
\usepackage{fancybox,framed}
\usepackage{arydshln}
\usepackage{xcolor}
\usepackage{algorithm,algorithmic}
\usepackage{pdfpages}

\usepackage{fmtcount}
\makeatletter
\renewcommand{\@makechapterhead}[1]{\vspace *{40\p@ }{\parindent \z@ 
\raggedright \normalfont \ifnum \c@secnumdepth >\m@ne \Huge \bfseries 
\@chapapp \space \Numberstring{chapter} \vskip 10\p@ \fi #1\par \nobreak \vskip 30\p@ }}
\makeatother

\newcommand{\argmax}{\mathop{\rm argmax}\limits}

\newcommand{\softmax}{\mathop{\rm softmax}\limits}

\DeclareFontFamily{OMS}{oasy}{\skewchar\font48 }
\DeclareFontShape{OMS}{oasy}{m}{n}{%
         <-5.5> oasy5     <5.5-6.5> oasy6
      <6.5-7.5> oasy7     <7.5-8.5> oasy8
      <8.5-9.5> oasy9     <9.5->  oasy10
      }{}
\DeclareFontShape{OMS}{oasy}{b}{n}{%
       <-6> oabsy5
      <6-8> oabsy7
      <8->  oabsy10
      }{}
\DeclareSymbolFont{oasy}{OMS}{oasy}{m}{n}
\SetSymbolFont{oasy}{bold}{OMS}{oasy}{b}{n}

\DeclareMathSymbol{\smallleftarrow}     {\mathrel}{oasy}{"20}
\DeclareMathSymbol{\smallrightarrow}    {\mathrel}{oasy}{"21}
\DeclareMathSymbol{\smallleftrightarrow}{\mathrel}{oasy}{"24}

\newenvironment{itemizesquish}{\begin{list}{\labelitemi}{\setlength{\itemsep}{-0.2em}\setlength{\labelwidth}{0.5em}\setlength{\leftmargin}{\labelwidth}\addtolength{\leftmargin}{\labelsep}}}{\end{list}}
\begin{document}
\hypersetup{breaklinks=true}

 \frontmatter

 \makepreliminarypages
 \newpage

 \doublespace
 \addcontentsline{toc}{part}{LIST OF CONTENTS}
 \tableofcontents
 \thispagestyle{plain}

\renewcommand{\cfttabfont}{Table }
 \mylistoftables
 \thispagestyle{plain}
 
  \renewcommand{\cftfigfont}{Figure }
 \mylistoffigures
 \thispagestyle{plain}

 \clearemptydoublepage

 \mainmatter

\addchapheadtotoc
\chapter{Introduction}\label{sec:introduction}

In many areas of engineering, it is our dream to create a machine that is more productive than a human, and can tolerate working for a longer time and that releases workers from tedious tasks. In Artificial Intelligence (AI) we seek a machine that has intelligence of humans. Here, intelligence might include the ability to understand images, understand speech, and read texts. 

The ability to read texts is studied in Natural Language Processing (NLP), a field of study to process natural language texts, and its ultimate goal is to create a machine that understands natural language texts. Although the ability is essential for the desired machine to communicate with humans like workers do, understanding texts is not a well-defined goal, and it is nontrivial to verify the ability. 

A legacy approach for verifying the ability is the Turing test~\citep{10.1093/mind/LIX.236.433} where a tester talks with a machine or human, and we see whether the tester can reliably tell the machine from the human or not. Although the test setting is convincing, there are two practical issues that we are concerned about. First, the test cannot compare the intelligence of two given machines. The test verifies if each machine has the intelligence or not and each machine makes a fairly independent conversation for each other, thus it is difficult to compare these test results. Second, the test only verifies the existence of the intelligence, and it does not help to explain how the machine understands given texts.

A practical approach might be reading comprehension tasks, where machines answer a question about a given passage rather than making a conversation. Here it is important that answering the question requires information described in the passage. So we can see how much a machine understands the given passage by observing the answer the machine makes. In this setting, we can compare the abilities of these machines by simply counting the number of correct answers given by each machine. 

Additionally, we are also interested in how the information in texts is represented in the machines, especially deep neural network models that are notoriously difficult to interpret. We challenge this question with focusing entities and their relations described in the texts and show here that the vectors of neural readers can be decomposed into a predicate and entities.

Thus, this dissertation shows studies of these reading comprehension tasks focusing on entities and relations. We believe that understanding how machines take care of entities and their relations in a given passage helps further the study of machine reading comprehension. Then eventually, this study contributes to the ultimate goals of AI.

\section{Reading Comprehension}
\begin{figure}[t]
    \begin{center}
        \includegraphics[width=0.8\textwidth]{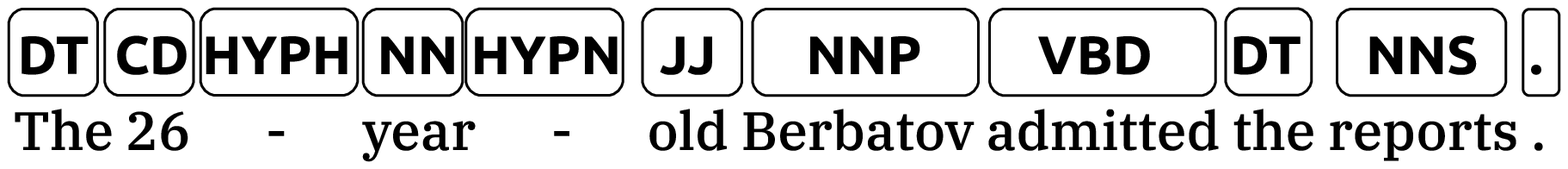}
        \caption{An example of part-of-speech tagging. Each tag indicates a part-of-speech of each token; DT (determiner), CD (cardinal number), HYPH (hyphen), JJ (adjective), NNP (proper noun, singular), VBD (verb, past tense), and NNS (noun, plural). }
        \label{fig:example_pos}
    
    \end{center}
        \begin{center}
        \includegraphics[width=0.8\textwidth]{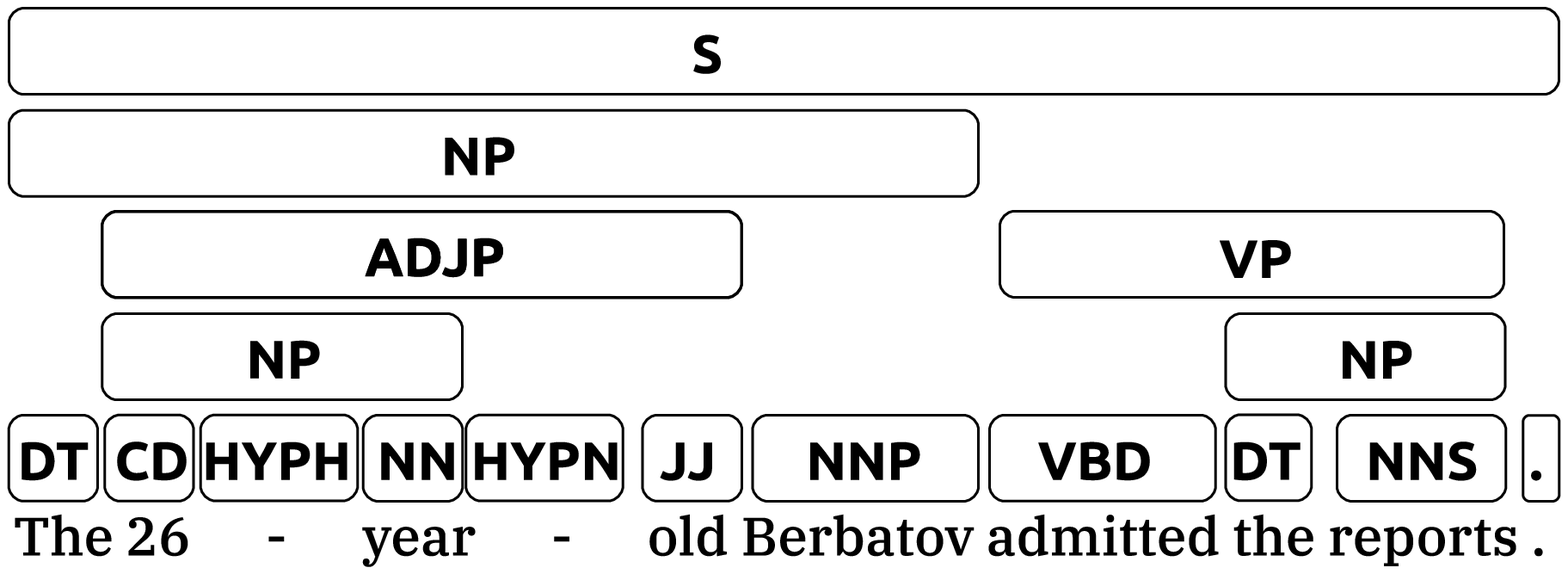}
        \caption{An example of syntactic parsing. Each tag indicates a type of phrase; VP (verb phrase), NP (noun phrase), ADJP (adjective phrase.), S (sentence)}
        \label{fig:example_syntactic_parse}
    \end{center}
\end{figure}

\begin{figure}[t]
    \begin{center}
        \includegraphics[width=0.7\textwidth]{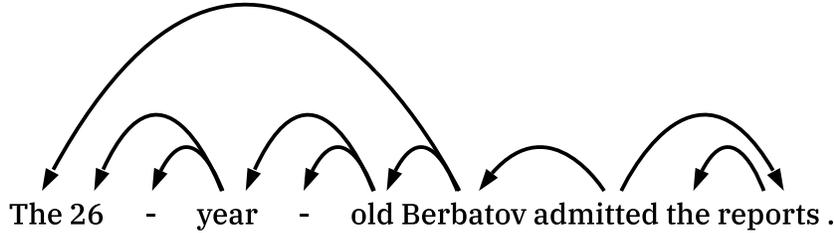}
        \caption{An example of dependency parsing.}
        \label{fig:example_dependency}
    \end{center}
\end{figure}

A machine that understands human language is the ultimate goal of NLP. Understanding is a nontrivial concept to define; however, the NLP community believes it involves multiple aspects and has put decades of effort into solving different tasks for the various aspects of text understanding, including:

\textbf{Syntactic aspects:}
\begin{itemize}
\item Part-of-speech tagging: This is a task to find a syntactic rule for each token in a sentence. Each token is identified as a noun, verb, adjective, etc. Figure \ref{fig:example_pos} shows an example of part-of-speech tagging. 

\item Syntactic parsing: This is a task to find syntactic phrases in a sentence such as a noun phrase, verb phrase. Figure \ref{fig:example_syntactic_parse} shows an example of syntactic parsing. 

\item Dependency parsing: Dependency is a relation between tokens where a token modifies another token. Dependency parsing is a task to find all dependencies in a sentence. Figure \ref{fig:example_dependency} shows an example of dependency parsing.  

\end{itemize}

\begin{figure}[t]
    \begin{center}
        \includegraphics[width=0.7\textwidth]{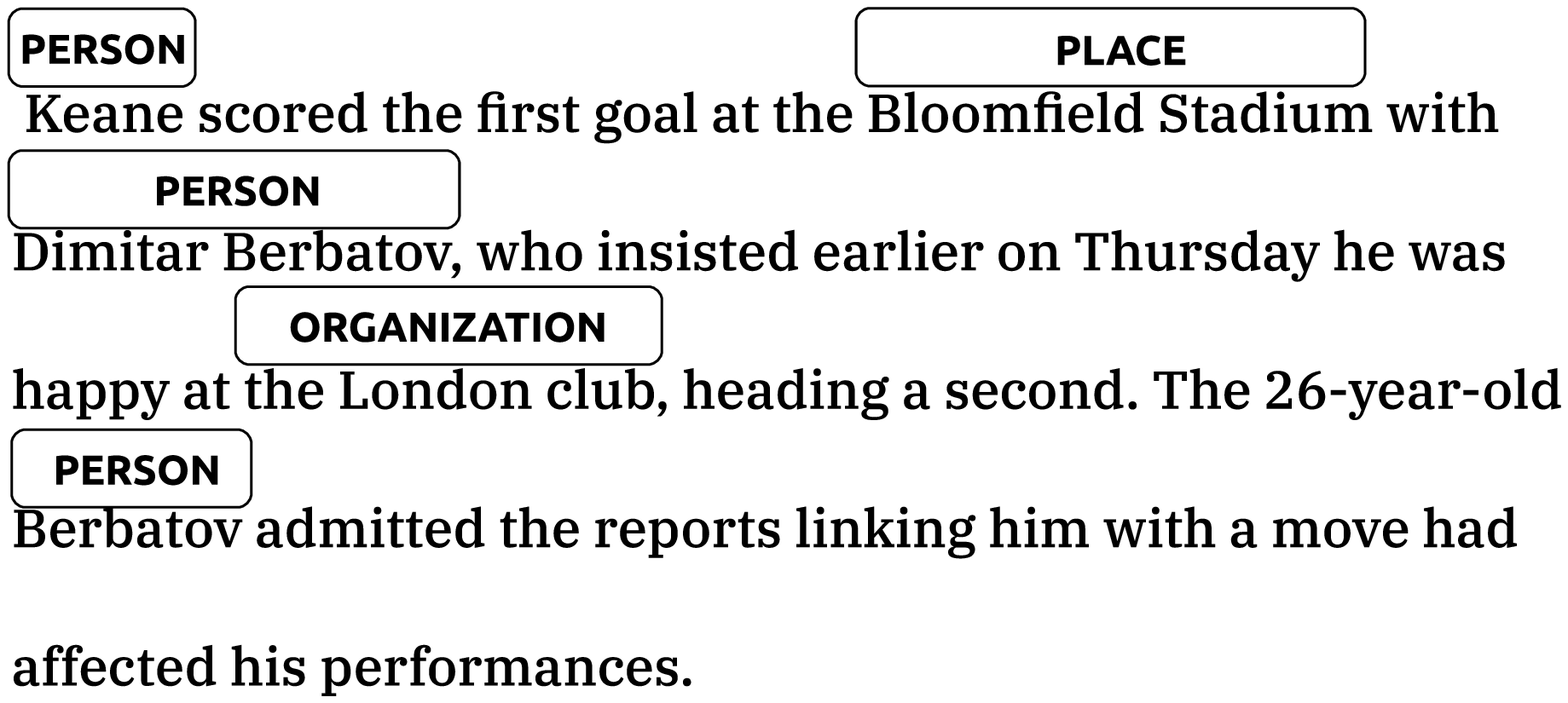}
        \caption{An example of named entity recognition. Here, person names, place names, and organization names are recognized.}
        \label{fig:example_ner}
    \end{center}

    \begin{center}
        \includegraphics[width=0.7\textwidth]{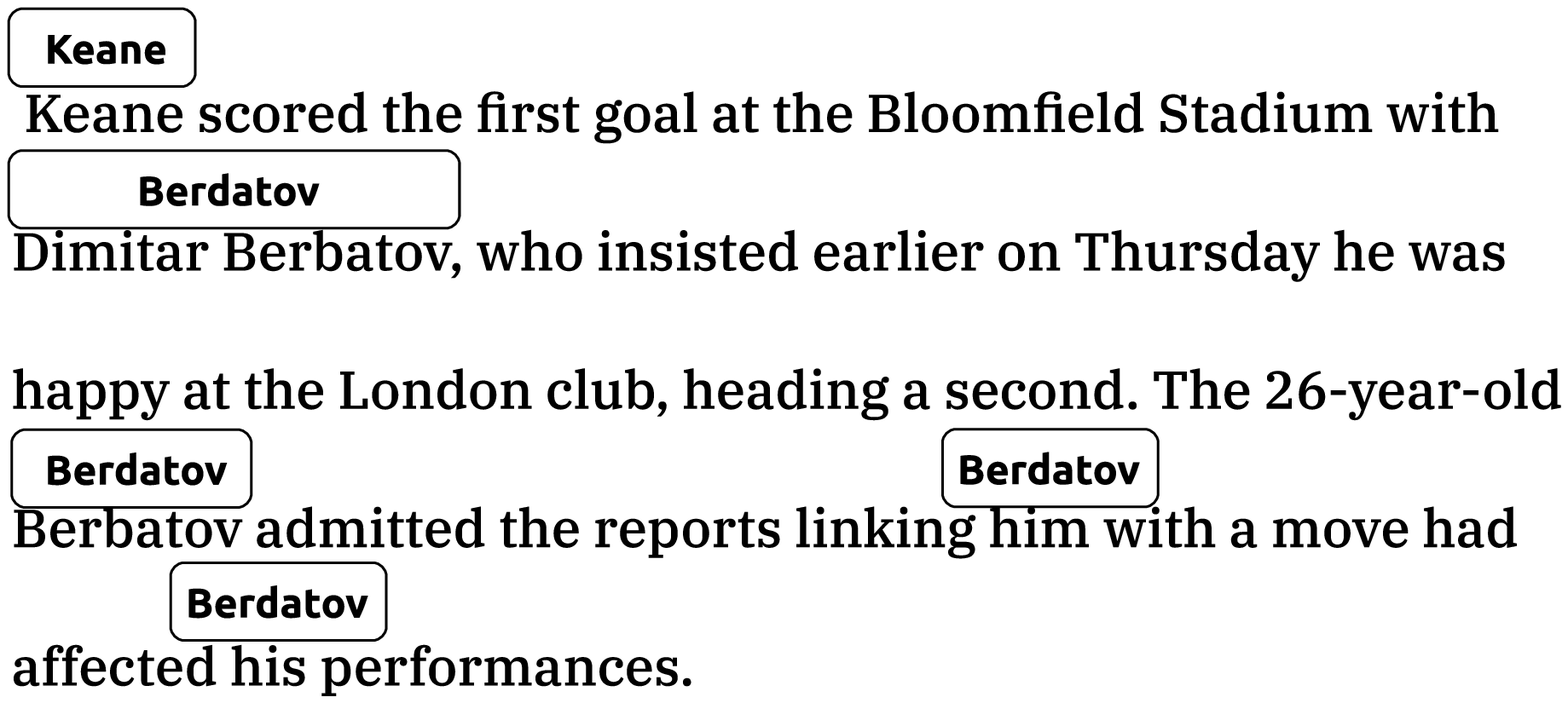}
        \caption{An example of coreference resolution. Here, two person entities; Robbie Keane and Dimitar Berbatov, are recognized.}
        \label{fig:example_coref}
    \end{center}
\end{figure}

\textbf{Semantic aspects:}
\begin{itemize}

\item Named entity recognition: This is a task to find named entities and their types in a sentence. Typical named entity types are ``Person'' and ``Location''. Figure \ref{fig:example_ner} shows an example of named entity recognition. 

\item Coreference resolution: This is a task to collect tokens that refer to the same entity. For example, Donald Trump can be referred by ``he'', ``Trump'' or ``the president.'' Figure \ref{fig:example_coref} shows an example of coreference resolution. 

\end{itemize}

A reading comprehension task is a question answering task that is designed for testing all these aspects and probe even deeper levels of understanding. Table \ref{table:rc_example} shows an example of a reading comprehension question from Who-did-What~\cite{Onishi2016}. Here a machine selects the most appropriate answer to fill the blank in the question from the choice list.  To solve the question, the machine needs to understand syntax, including the part-of-speech tags of each token, syntactic and dependency structures; thus, it finds tokens referring candidate answers; (1) Robbie Keane and (2) Dimitar Berbato, in the passage with named entity recognition and coreference resolution, and then it might find ``Dimitar Berbato'' is the best answer.

\begin{table*}[t]
	\centering
	\fbox{\begin{minipage}[t]{\textwidth}
{\footnotesize
{\bf Passage:} 
Tottenham won 2-0 at Hapoel Tel Aviv in UEFA Cup action on Thursday night in a defensive display which impressed Spurs skipper Robbie Keane. ...  Keane scored the first goal at the Bloomfield Stadium with Dimitar Berbatov, who insisted earlier on Thursday he was happy at the London club, heading a second. The 26-year-old Berbatov admitted the reports linking him with a move had affected his performances ...  Spurs manager Juande Ramos has won the UEFA Cup in the last two seasons ...

\vspace{1ex}
{\bf Question:} Tottenham manager Juande Ramos has hinted he will allow *** to leave if the Bulgaria striker makes it clear he is unhappy.

\vspace{1ex}
{\bf Choices:} (1) Robbie Keane (2) Dimitar Berbatov
}
\end{minipage}}
\caption{An example of reading comprehension question.}
\label{table:rc_example}
\end{table*}

\subsection{Problem Formulations}
Multiple reading comprehension tasks with different styles have been studied (see examples in Section \ref{sec:wdw:related}). In these reading comprehension tasks, a machine takes a passage and question then returns an answer. Hence, a supervised training instance is a tuple of a passage, question, and answer. The passage is a text resource that provides enough information to find the answer, such as a news article, encyclopedia article, or multiple paragraphs of these articles. The question is also a text resource, but it is much shorter than the passage. The answer style is different depending on the style of each reading comprehension task. Here, we divide existing reading comprehension tasks into three styles depending on their answer type.

\begin{itemize}
\item Multiple choice: In this style, a list of candidate answers is given along with each passage and question. Hence the answer is one of the candidate answers. On the example question in Table \ref{table:intro.qa}, the candidate answers are all viruses mentioned in the passage, and the correct answer is \textit{(a)COVID-19}. Each dataset has a different algorithm to pick these candidate answers. For example, bAbI~\citep{Weston2015} picked all nouns in the passage for the candidate answers, candidate answers in CNN/Daily Mail dataset~\citep{Hermann2015} are all entities in the passage, and candidate answers in WDW~\citep{Onishi2016} are a subset of person names in the passage (details in Chapter \ref{sec:work1}).  \\
The performance of a machine is evaluated by the accuracy; the number of correct answers over the number of all questions. 

\item Span prediction: In this style, the answer is a span in the passage, i.e., the answer is a pair of a start token and end token. This style is also referred to as extractive question answering. On the example question in Table \ref{table:intro.qa}, there are two occurrences of \textit{COVID-19} in the passage, but the answer is the second one. \\
The performance of a machine is evaluated by span-level accuracy by exact matching (EM) and/or an F1 score. EM is the same as the accuracy where the predicted span is correct if and only if the sequence of words specified by the predicated span is the same as the sequence of words specified by the gold span. This matching scheme might be called string matching.  The F1 score is a harmonic mean of precision and recall that are computed between the bag of tokens in the predicted span and the bag of tokens in the gold span. 
\begin{eqnarray}
    \textrm{Precision} &=& \frac{|P \cap G|}{|P|}, \\
    \textrm{Recall} &=& \frac{|P \cap G|}{|G|}, \\ \nonumber
\end{eqnarray}
\begin{equation}
    \textrm{F1} = \frac{2 \times \textrm{Precision} \times \textrm{Recall}}{\textrm{Precision} + \textrm{Recall}},
\end{equation}
where $P$ and $G$ are the bag of tokens in the predicted span and that in the gold span, respectively. 

\item Free-form answer: In this style, the answer can be any sequence of words in a vocabulary; thus, a machine generates the sequence to answer the given question. On the example question in Table \ref{table:intro.qa}, the answer is ``COVID-19'' (string). The evaluation is not trivial and different for each dataset of this style. \\
Wikireading~\citep{wikireading2016} employs EM and F1, others employ standard metrics for natural language generation tasks including Bilingual Evaluation Understudy (BLEU)~\citep{papineni-etal-2002-bleu}, Meteor~\citep{banerjee-lavie-2005-meteor} and Recall-Oriented Understudy for Gisting Evaluation (ROUGE)~\citep{lin-2004-rouge}. 

\end{itemize}

\begin{table}[th!]
	\centering
	\fbox{\begin{minipage}[t]{\textwidth}

\textbf{Passage:} Pregnant women may be at higher risk for severe infection with COVID-19 based on data from other similar viruses, like SARS and MERS, but data for $\rm \underset{\textbf{Answer[span]}}{COVID-19}$ is lacking.

\textbf{Question:} We are lacking for the data of \#BLANK\# to evaluate the risk of pregnant woman.

\textbf{Candidate answers:} a)COVID-19, b)SARS, c)MERS

\textbf{Answer[ID]:} (a)

\textbf{Answer[free-form]:} COVID-19

    \end{minipage}}
    \caption{Reading comprehension question answering whose answer is an entity. Answer[ID] is an answer selected from the candidate answers. Answer[span] is an answer identified by a span.}
    \label{table:intro.qa}
\end{table}

\subsection{Reading comprehension task and other question answering tasks} \label{sec:rc_vs_qa}
Reading comprehension tasks are closely related to other question answering tasks because they are essentially question answering problems over a passage, a relatively short text. Thus, reading comprehension tasks and other question answering tasks share many common characteristics in their problem formulation, approaches and evaluation. However, it is worth noting that the goal of reading comprehension tasks is different from the goal of other question answering tasks. 

The goal of other question answering tasks is to appropriately answer questions posed by humans, and reading comprehension skills are less considered. Thus the machine may use any kind of information resources, including structured knowledge such as knowledge bases and unstructured knowledge texts such as encyclopedias, dictionaries, news articles, and Web texts. Additionally, the unstructured knowledge texts are longer than a passage and typically web-scale. These information resources require less reading comprehension skills described in Chapter \ref{sec:introduction}. For example, given an access to a large text corpus, a simple grammatical transformation and string matching will likely suffice to answer the question like ``who is the president of the U.S.'' Here the question can be grammatically transformed into a declarative sentence, ``*** is the president of the U.S.'' Then, the machine more likely finds a sentence that matches the declarative sentence.

On the other hand, the goal of reading comprehension is to understand a given (short) text. Thus a machine uses unstructured knowledge texts only. The texts are typically short and carefully written so that they require more reading comprehension skills. For example, multiple given passages might share some information. Such shared information is called world knowledge, and some machines might be able to answer a question correctly without reading the given passage but using the world knowledge written in other passages. Hence, this issue makes it difficult to tell if the machine has reading comprehension skills. To avoid this issue, early work in this field mostly focused on fictional stories~\citep{10.5555/1624435.1624467} because each fictional story has different characters and stories and then unlikely shares information.

Early study~\citep{Matt2009} describes this difference by using terms; micro-reading/macro-reading. Macro-reading is a task where the input is a large text collection, and the output is a large collection of facts expressed by the text collection, without requiring that every fact be extracted. Micro-reading is a task where a single text document is input, and the desired output is the full information content of that document. 


\subsection{History}
Reading comprehension question answering is not new, and we can find early work from the 1970s. In this section, we review the history of three paradigms: development of the theory, rule-based systems, and deep learning systems. 

Very early systems operate in very limited domains in the 1970s. For example, SHRDLU~\citep{winograd1971procedures} is a computer program where a user can move some objects in a 3D computer graphic by using English. LUNAR~\citep{woods1972lunar} is another computer program that answers questions about lunar geology and chemistry, and Baseball~\citep{green1961baseball} is for questions about baseball. 

One of the most notable early work in the 1970s might be the QUALM system~\citep{10.5555/1624435.1624467}. The work proposed a conceptual theory to understand the nature of question answering. Here the work analyzed how humans classify questions, and the algorithm classified questions in a similar way that humans do.

In the 1980s to 1990s, various rule-based systems were proposed for each domain. Here we describe a notable shared task and dataset. The dataset was proposed by \citet{hirschman-etal-1999-deep} and consists of 60 stories for development and 60 stories for testing of 3rd to 6th grade material, and each story is followed by short-answer questions, i.e., who, what, when, where and why questions. In the task, a machine takes each story and question then finds a sentence in the story that most likely contains the answer key. Multiple rule-based systems were developed for this task. Deep Read~\citep{hirschman-etal-1999-deep} takes a bag-of-words approach with shallow linguistic processing, including stemming, name identification, semantic
class identification, and pronoun resolution. QUARC~\citep{Riloff2000ARQ} uses lexical and semantic correspondence, and then ~\citet{10.3115/1117595.1117596} combines them. As the results, these systems achieved 30–40\% accuracy, i.e., these systems correctly predict a sentence containing the answer for 30-40\% of questions.

From the 2010s, supervised learning models significantly improved their performance in various tasks, including reading comprehension tasks. Even some supervised learning models overcame human performance in some tasks~\citep{chen2016}. These improvements were made by deep neural networks and large-scale datasets. 

A deep neural network is a scalable machine learning model. A deep neural network is typically composed of ``units''. Each unit takes an input vector $x$ and returns an output vector $y$ by using a linear and non-linear transformation as the following.
\begin{equation}
    y = f( W \cdot x + b ),
\end{equation}
where $W$ is a matrix, $b$ is a bias vector, and $f$ is a non-linear function. 
The deep neural network is trained by a stochastic gradient descent algorithm where a loss is computed on a subset of training instances, and then the gradient of the loss is computed with respecting the parameters of the deep neural network. Hence, the parameters are updated to the direction of the gradient. 
\begin{eqnarray}
    \mathcal{L}_{mini}^\theta(X') = \sum_{x \in X'} \mathcal{L}^\theta(x), \\ 
    \theta \Leftarrow \theta - \lambda \frac{\partial \mathcal{L}_{mini}^\theta(X)}{\partial \mathcal{\theta}},
\end{eqnarray}
where $\mathcal{L}$ is the loss function to be minimized, $X'$ is a subset of training instances called mini-batch, and $\theta$ is the parameters. 
The stochastic gradient algorithm takes linear time against the size of the training data, and the memory requirement is linear to the size of the mini-batch. Thus neural network models can learn any large-scale training data in linear time by the stochastic gradient algorithm.

Larger training data provides more instances to learn, hence scaling up training data is believed to be a promising approach in machine learning. Here, we note the contribution of the World Wide Web (WWW) to the large-scale training data. The WWW is an information system over the Internet where a document or web resource is identified by a Uniform Resource Locators (URL). People uploads various kinds of texts on the WWW, including news articles, blog articles, and encyclopedia articles. The amount of these texts on the WWW was estimated as at least 320 million pages in 1998~\citep{Lawrence98}, and it is estimated as at least billions in 2016~\citep{bosch2016}. Naturally, these texts are computer-readable, unlike texts on books, and some of them are copy-right free. Hence we find the text on the WWW as a large accessible text resource. Recently, the WWW is a major resource of multiple standard reading comprehension datasets including, SQuAD, Wikihop, HotpotQA~\citep{Rajpurkar2016SQuAD:Text, welbl-etal-2018-constructing, yang-etal-2018-hotpotqa}. 

Thanks to the large-scale dataset supported by the WWW and the scalable training algorithm, deep neural network models can learn significantly large information on the dataset. As a result, these models perform better and better, and then their performances are achieving the human performance in some tasks~\citep{chen2016}. 

The significant success of deep learning raises two questions. 
\begin{itemize}
    \item ``What is a good question in reading comprehension tasks?''
     \item ``How do these machines understand texts?''
\end{itemize}
Questions in reading comprehension tasks are designed for testing reading comprehension skills, and each question requires these skills to solve. Today, as the deep neural network models perform better and better, we are more and more interested in more complicated reading comprehension skills that are beyond NER, coreference resolution, and dependency parsing. Additionally, we need to feed millions of such questions to train the deep neural network models, and it is not realistic for us to write each question manually. To address the problem and provide millions of such comprehension questions, we take a sampling approach in Chapter \ref{sec:work1}.

Early systems were rule-based, and the mechanism of their text-understanding is relatively explainable. For example, if a machine reads a given text by operating rules designed by a researcher, then the process can be explained by a sequence of rules that the machine used. This sequence explains how the machine understands the given texts. On the other hand, deep neural network models operate multiple vector transformations, and each transformation does not explicitly correlate with any grammatical/semantic rules. Thus, unlike rule-based systems, the sequence of these operations does not explain enough how the machine understands the given text. We claim that entities and their relations can be a key to explainability in Section \ref{sec:entity_relation}. Then, we empirically analyze how neural network models understand texts by using entities and their relations in Chapter \ref{sec:work2}, and apply it to our novel neural reader in Chapter \ref{sec:wikihop}. In Chapter \ref{sec:work3}, we extracted these entities and relations and visualize them for materials science.

\section{Entity and Relation} \label{sec:entity_relation}

We are interested in entities and their relations in the context of reading comprehension. In the following, we overview entities and their relations in the context of knowledge bases. Then, we describe reading comprehension datasets focusing on entities and relations, and also relation extraction from the point of view of reading comprehension. 

\begin{figure}[t]
    \begin{center}
        \includegraphics[width=0.9\textwidth]{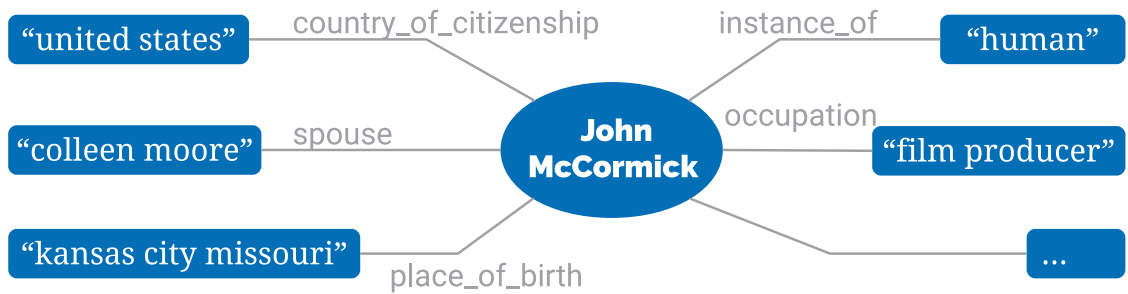}
        \caption{Entities and their relations around ``John McCormick'' in Wikidata.}
        \label{fig:sample_realtions}
    \end{center}
\end{figure}

Entities and their relations are well studied in the context of knowledge bases. A knowledge base such as WordNet~\citep{Fellbaum1998} or Wikidata~\citep{Vrandecic:2014:WFC:2661061.2629489} is a structured database that typically represents its information by using entities and their relations as Fig.\ref{fig:sample_realtions} shows the relations around ``John McCormick''.
Here, entities and their relation are defined for the information desired to be represented.  \citet{quine1948there} stated that ``To be assumed as an entity is [...] to be reckoned as the value of a variable'' or ``to be is to be the value of a variable
''. \citet{hobbs1985ontological}, inspired by Quine, limited entity types to ``physical object, numbers, sets, times, possible worlds, propositions, events''. Naturally, their relations are also designed for the target information.

Entities and their relations are critical to solve questions in some reading comprehension question answering tasks. For example, each answer of CNN/Daily Mail dataset~\citep{Hermann2015} is an entity that satisfies the condition given by the question sentence. The dataset is Cloze-style, where each question is a sentence whose key entity is blanked out. Here the question asks to find the blanked entity from the given passage. In other cases, each question of Wikireading~\citep{wikireading2016} and Wikihop~\citep{welbl-etal-2018-constructing} consists of an entity and/or relation. In Wikihop, each question is a pair of a subject entity and relation, and the answer is an object entity that has the relation with the subject entity. In Wikireading, each question is a relation and the passage describes a subject entity and the answer is an object entity that has the relation with the subject entity described in the passage. 

We also consider question answering tasks whose answers are relations. These tasks are studied in the context of relation extraction in knowledge base population described in Section \ref{sec:intro.kbp}. Relation extraction is a task for finding a relation between two given entities described in a text resource. It is worth noting that the task is different from relation classification. Relation classification is a task for finding a relation between two given entities described in a given text resource (typically a sentence) where the positions of these entities are given. On the other hand, the positions are not given in relation extraction, and the text resource is typically longer than a single sentence. Thus, the task can be viewed as another reading comprehension task focusing entities and relations in the text. 

Entities and relations are critical for these tasks; however, we believe that such entities and their relations are critical, not only for these datasets but also for other datasets that implicitly require a machine to understand entities and their relations.

\subsection{Knowledge base population} \label{sec:intro.kbp}

In this section, we briefly overview how knowledge bases help various tasks, including question answering and information retrieval, and the motivation of knowledge base population, a task to fill a knowledge base from texts.

A knowledge base is often a critical component of an expert system. An expert system is typically composed of inference rules written by hand and a knowledge base and emulates the decision-making ability of a human expert. As it is sometimes difficult for the human expert to explain his/her decision, it is difficult to design complicated inference rules, but it might be easier to add more knowledge to the knowledge base. The performance of each system heavily depends on the coverage of its knowledge base.

Today, some large-scale knowledge bases are available, e.g., Freebase and Wikidata. Freebase started as a collaborative knowledge base whose data was accumulated by its community members.
Freebase consists of 125M tuples of a subject entity, object entity, and their relation, whose topics spread over 4K types, including people, media, and locations~\citep{bollacker2008freeabse, freebase:datadumps}. Wikidata is also a collaborative knowledge base consisting of 87M entities\footnote{https://www.wikidata.org/wiki/Wikidata:Main\_Page (last accessed in June 2020)} and most of these entities are linked to entities in sister projects such as Wikipedia; thus, it can provide extra information about these entities. 
Such large-scale knowledge bases help various tasks, including information retrieval and question answering, but still, the coverage of the knowledge base is critical for the performance.

Despite the efforts of the community members who are maintaining these knowledge bases, their sizes are far from sufficient because new knowledge is emerging rapidly. On the other hand, we are more likely able to access textual information describing the new knowledge. Thus, we study knowledge base population to feed the knowledge base from texts.



\chapter{Entity-centered reading comprehension dataset} \label{sec:work1}


\label{sec:wdw:intro}
Researchers distinguish the problem of general knowledge question answering from that of reading comprehension~\citep{Hermann2015, Hill2016TheRepresentations} as descibed in Section \ref{sec:rc_vs_qa}. Reading comprehension is more difficult than knowledge-based or Information Retrieval (IR)-based question answering in two ways. First, reading comprehension systems must infer answers from a given unstructured passage rather than structured knowledge sources such as Freebase~\cite{bollacker2008freeabse} or the Google Knowledge Graph~\cite{ Singhal2012IntroducingStrings}. Second, reading comprehension systems cannot exploit the large level of redundancy present on the web to find statements that provide a strong syntactic match to the question~\cite{Yang2015}. In contrast, a reading comprehension system must use the single phrasing in the given passage, which may be a poor syntactic match to the question.

\begin{table*}[th!]
	\centering
	\fbox{\begin{minipage}[t]{\textwidth}
{\footnotesize
{\bf Passage:}
Britain's decision on Thursday to drop extradition proceedings against
Gen.~Augusto Pinochet and allow him to return to Chile is
understandably frustrating ... Jack Straw, the home
secretary, said the 84-year-old former dictator's ability to
understand the charges against him and to direct his defense had been
seriously impaired by a series of strokes. ... Chile's
president-elect, Ricardo Lagos, has wisely pledged to let justice run
its course. But the outgoing government of President Eduardo Frei is
pushing a constitutional reform that would allow Pinochet to step down
from the Senate and retain parliamentary immunity from
prosecution. ...

\vspace{1ex}
{\bf Question:} 
Sources close to the presidential palace said that Fujimori declined at the last moment to leave the country and instead he will send a high level delegation to the ceremony, at which Chilean President Eduardo Frei will pass the mandate to ***.

\vspace{1ex}
{\bf Choices:} (1) Augusto Pinochet (2) Jack Straw (3) Ricardo Lagos

\vspace{2ex}
{\bf Passage:} 
Tottenham won 2-0 at Hapoel Tel Aviv in UEFA Cup action on Thursday
night in a defensive display which impressed Spurs skipper Robbie
Keane. ...  Keane scored the first goal at the Bloomfield Stadium with Dimitar Berbatov, who insisted earlier on Thursday he was happy at the London club, heading a second.
The
26-year-old Berbatov admitted the reports linking him with a move had
affected his performances ...  Spurs manager
Juande Ramos has won the UEFA Cup in the last two seasons ...

\vspace{1ex}
{\bf Question:} Tottenham manager Juande Ramos has hinted he will allow *** to leave if the Bulgaria striker makes it clear he is unhappy.

\vspace{1ex}
{\bf Choices:} (1) Robbie Keane (2) Dimitar Berbatov
}
\end{minipage}}
\caption{Sample reading comprehension problems from our dataset.}
\label{table:wdw_sample}
\end{table*}

In this chapter, we describe the construction of a new reading comprehension dataset that we refer to as Who-did-What (WDW)~\citep{Onishi2016}. Two typical examples are shown in Table~\ref{table:wdw_sample}.\footnote{The passages here only show certain salient portions of the passage. In the actual dataset, the entire article is given. The correct answers are (3) and (2).} The process of forming a problem starts with the selection of a question article from the English Gigaword corpus. The question is formed by deleting a person named entity from the first sentence of the question article. An information retrieval system is then used to select a passage with high overlap with the first sentence of the question article, and an answer choice list is generated from the person named entities in the passage.

Our dataset differs from the CNN/Daily Mail dataset~\citep{Hermann2015} in that it forms questions from two distinct articles rather than summary points. This allows problems to be derived from document collections that do not contain manually-written summaries. This also reduces the syntactic similarity between the question and the relevant sentences in the passage, increasing the need for deeper semantic analysis.

To make the dataset more challenging we selectively remove problems so as to suppress four simple baselines --- selecting the most mentioned person, the first mentioned person, and two language model baselines. This is also intended to produce problems requiring deeper semantic analysis.

The resulting dataset yields a larger gap between human and machine performance than existing ones. Humans can answer questions in our dataset with an 84\% success rate compared to the estimates of 75\% for CNN~\citep{chen2016} and 82\% for the CBT named entities task \citep{Hill2016TheRepresentations}.  In spite of this higher level of human performance, various existing readers perform significantly worse on our dataset than they do on the CNN dataset.  For example, the Attentive Reader \cite{Hermann2015} achieves 63\% on CNN but only 55\% on WDW and the Attention Sum Reader~\cite{Kadlec2016} achieves 70\% on CNN but only 59\% on WDW.

In summary, we believe that our WDW is more challenging, and requires deeper semantic analysis.

\section{Related work}
\label{sec:wdw:related}
Our WDW is related to several datasets for machine comprehension. In this section, we review notable reading comprehension datasets since the 1990s including dataset developed after our WDW. 

The Deep Read dataset~\citep{hirschman-etal-1999-deep} is an outstanding early work on reading comprehension dataset. The dataset consists of 60 development and 60 test simulated news stories of 3rd to 6th grade material. Each story is followed by short-answer 5W questions; who, what, when, where, and why questions, as a sample on Table \ref{table:deepread}. These stories and questions are entirely hand-written. The dataset is significantly smaller than other datasets, i.e., 60 stories $\times$ 5 questions. Hence, it is difficult to apply machine learning models with a large number of parameters. 

\begin{table*}[t!]
	\centering
	\fbox{\begin{minipage}[t]{\textwidth}

\textbf{Passage:} \\
Library of Congress Has Books for Everyone (WASHINGTON, D.C., 1964) - It was 150 years ago this year that our nation's biggest library burned to the ground. Copies of all the wriuen books of the time were kept in the Library of Congress. But they were destroyed by fire in 1814 during a war with the British. That fire didn't stop book lovers. The next year, they began to rebuild the library. By giving it 6,457 of his books, Thomas Jefferson helped get it started. The first libraries in the United States could be used by members only. But the Library of Congress was built for all the people. From the start, it was our national library. Today, the Library of Congress is one of the largest libraries in the world. People can find a copy of just about every book and magazine printed. Libraries have been with us since people first learned to write. One of the oldest to be found dates back to about 800 years B.C. The books were written on tablets made from clay. The people who took care of the books were called ``men of the written tablets.''

\textbf{Question1:} Who gave books to the new library?

\textbf{Question2:} What is the name of our national library?

\textbf{Question3:} When did this library burn down? 

\textbf{Question4:} Where can this library be found? 

\textbf{Question5:} Why were some early people called ``men of the written tablets''?

    \end{minipage}}
    \caption{A sample question from Remedia Reading Comprehension Story and Questions. }
    \label{table:deepread}
\end{table*}

The MCTest dataset~\citep{Richardson2013} consists of 660 fictional stories with four multiple choice questions each. A sample is given in Table \ref{table:mctest.sample}. Each question is expected to be answerable by seven year old children. These fictional stories and questions were written by Amazon Mechanical Turk cloud workers. Although they claim that their cloud sourcing approach is scalable, this dataset is too small to train models for the general problem of reading comprehension. 

\begin{table*}[t!]
	\centering
	\fbox{\begin{minipage}[t]{\textwidth}

\textbf{Passage:} James the Turtle was always getting in trouble. Sometimes he'd reach into the freezer and empty out all the food. Other times he'd sled on the deck and get a splinter. His aunt Jane tried as hard as she could to keep him out of trouble, but he was sneaky and got into lots of trouble behind her back. One day, James thought he would go into town and see what kind of trouble he could get into. He went to the grocery store and pulled all the pudding off the shelves and ate two jars. Then he walked to the fast food restaurant and ordered 15 bags of fries. He didn't pay, and instead headed home. His aunt was waiting for him in his room. She told James that she loved him, but he would have to start acting like a well-behaved turtle. After about a month, and after getting into lots of trouble, James finally made up his mind to be a better turtle.

\textbf{Question1:} What is the name of the trouble making turtle?

\textbf{Candidate answers:} a)Fries, b)Pudding, c)James, d)Jane

\textbf{Answer1:} (c)James

\textbf{Question2:} What did James pull off of the shelves in the grocery store?

\textbf{Candidate answers:} a)pudding, b)fries, c)food, d)splinters

\textbf{Answer2:} (a)pudding

    \end{minipage}}
    \caption{A sample question from MCTest. }
    \label{table:mctest.sample}
\end{table*}

The bAbI synthetic question answering dataset~\citep{Weston2015} contains passages describing a series of actions in a simulation followed by a question. For this synthetic data a logical algorithm can be written to solve the problems exactly (and, in fact, is used to generate ground truth answers). 

The Children's Book Test (CBT) dataset, created by Hill et al., contains 113,719 cloze-style named entity problems. Each problem consists of 20 consecutive sentences from a children's story, a 21st sentence in which a word has been deleted, and a list of ten choices for the deleted word, as a sample is given in Table \ref{table:cbt.sample}. The CBT dataset tests story completion rather than reading comprehension. The next event in a story is often not determined \---- surprises arise. This makes it difficult to predict the deleted word in the last sentence and may explain why human performance is lower for CBT than for our dataset. \---- 82\% for CBT vs.~84\% for WDW. The 16\% error rate for humans on WDW seems to be largely due to noise in problem formation introduced by errors in named entity recognition and parsing.  Reducing this noise in future versions of the dataset should significantly improve human performance. Another difference compared to CBT is that WDW has shorter choice lists on average. Random guessing achieves only 10\% on CBT but 32\% on WDW. The reduction in the number of choices seems likely to be responsible for the higher performance of an LSTM system on WDW \--- contextual LSTMs (the attentive reader of Hermann et al., 2015) improve from 44\% on CBT (as reported by Hill et al., 2016) to 55\% on WDW.

\begin{table*}[t!]
	\centering
	\fbox{\begin{minipage}[t]{\textwidth}

\textbf{Passage:} \\
1 Ring grew terribly afraid . \\
2 ` How do you like them ? ' \\
3 asked Snati . \\
4 ` Not well at all , ' said the Prince . \\
... \\
15 He came to the King and said he had something to say to him .\\
16 ` What is that ? ' \\
17 said the King . \\
18 Red said that he had just remembered the gold cloak , gold chess-board , and bright gold piece that the King had lost about a year before .\\
19 ` Do n't remind me of them ! ' \\
20 said the King . \\
21 Red , however , went on to say that , since Ring was such a mighty man that he could do everything , it had occurred to him to advise the \#BLANK\# to ask him to search for these treasures , and come back with them before Christmas ; in return the King should promise him his daughter .

\textbf{Candidate answers:} a)Dog, b)King, c)Prince, d)Red,...

\textbf{Answer:} King

    \end{minipage}}
    \caption{A sample question from the CBT dataset. }
    \label{table:cbt.sample}
\end{table*}

The CNN/Daily Mail datasets together consist of 1.4 million questions constructed from approximately 300,000 articles. Of existing datasets, these are the most similar to WDW in that they consist of cloze-style question answering problems derived from news articles.  Our WDW
differs from these datasets in not being derived from article summaries, in using baseline suppression, and in yielding a larger gap between machine and human performance.  WDW also differs in that the person named entities are not anonymized, permitting the use of external resources to improve performance while remaining difficult for language models due to suppression. 

\begin{table*}[t!]
	\centering
	\fbox{\begin{minipage}[t]{\textwidth}

\textbf{Passage:} ... a small aircraft carrying @entity5 , @entity6 and @entity7 the @entity12 @entity3 crashed a few miles from @entity9 , near @entity10 , @entity11 ...

\textbf{Question:} pilot error and snow were reasons stated for @placeholder plane crash

\textbf{Candidate answers:} 1)entity1, 2)entity2, 3)entity3, ...

\textbf{Answer[ID]:} (5)entity5

    \end{minipage}}
    \caption{A sample question from CNN/Daily Mail dataset. }
    \label{table:cnn_daily.sample}
\end{table*}

Stanford Question Answering Dataset (SQuAD)~\citep{Rajpurkar2016SQuAD:Text} contains more than 100K questions whose answer is a span of text in the given document. A sample question is given in Table \ref{table:squad.sample}. Questions and answer spans are written by cloud workers. In the dataset construction, a cloud worker writes five questions, and their answer spans for each passage that is a paragraph of a Wikipedia article whose length is shorter than 500 characters. In addition to the answer span, two other cloud workers are given the passage and question only and predict the answer span. Thus, each question has at most three gold answer spans. The evaluation metric is EM and F1. Here F1 is computed between a bag of tokens in a gold answer span and a bag of tokens in the predicted span. 

\begin{table*}[t!]
	\centering
	\fbox{\begin{minipage}[t]{\textwidth}

\textbf{Passage:} In meteorology, precipitation is any product of the condensation of atmospheric water vapor that falls under $\rm \underset{\textbf{answer1}}{gravity}$. The main forms of precipitation include drizzle, rain, sleet, snow, $\rm \underset{\rm \textbf{answer2}}{graupel}$ and hail... Precipitation forms as smaller droplets coalesce via collision with other rain drops or ice crystals $\rm \underset{\rm \textbf{answer3}}{within\ a\ cloud}$. Short, intense periods of rain in scattered locations are called “showers”...

\textbf{Question1:} What causes precipitation to fall?\\
\textbf{Question2:} What is another main form of precipitation besides drizzle, rain, snow, sleet and hail?\\
\textbf{Question3:} Where do water droplets collide with ice crystals to form precipitation?

    \end{minipage}}
    \caption{A sample question from SQuAD dataset. }
    \label{table:squad.sample}
\end{table*}

MS Machine Reading Comprehension (MS MARCO)~\citep{nguyen2016ms} is a reading comprehension dataset with the aspect of macro-reading. The dataset consists of 100K questions sampled from user queries issued to a search engine. Each question comes with a passage, which is a set of approximately ten web-pages that are retrieved by an information retrieval system. These questions and passages make the task more like a general question answering task rather than a reading comprehension task. Firstly, the passage is longer than that in other datasets whose passage is a paragraph or a news article. Secondly, it is unclear if answering these questions based on web-queries require the reading comprehension skills, e.g., we generally make a web-query by using keywords rather than a question sentence to help keyword matching. These aspects make these questions more likely to be solved by syntactic matching. 

TriviaQA~\citep{JoshiTriviaQA:Comprehension} is another reading comprehension dataset with the aspect of Macro-reading. The dataset consists of 96K questions and 663K evidence documents. These questions and their answers are from 14 trivial and quiz-league websites. The answer type is free-form answer, and the evaluation metrics are EM and F1 as following SQuAD. The evidence document is a passage in our context and collected from web-pages and Wikipedia articles by using a Web search engine. Hence, it is worth noting that each question has multiple evidence documents to read, unlike SQuAD where each question has one passage. Thus the passage is relatively long for each question, and then the dataset has the aspect of Macro-reading.

NarrativeQA~\citep{Kocisk2018TheChallenge} is a medium-scale reading comprehension dataset consisting of 1.5K passages and 47K questions. These questions are from books or movie scripts, and questions are written by cloud workers. In the dataset construction, the cloud workers write the pairs of a question and answer based solely on a given summary of the corresponding passage. The answer type is free-form answer, and then the evaluation metric is BLEU, Meteor and ROUGE, and the mean reciprocal rank (MRR). Here MRR is $\textrm{ave}_{r} \frac{1}{r}$ where $r$ is the rank of the correct answer among candidate answers. 

HotpotQA~\citep{yang-etal-2018-hotpotqa} is a reading comprehension dataset requiring the reasoning. Here the reasoning is a task to provide a set of sentences explaining why the answer is selected. The dataset consists of 113K questions and passages. Each passage is a set of paragraphs from Wikipedia articles, and the question is written by a cloud worker. Additionally, the cloud worker picks support facts, sentences in the passage that determine the answer for each question. The dataset employed Joint F1 for the evaluation metric in addition to EM and F1. Joint F1 is computed as follows:
\begin{equation}
	P^{(joint)} = P^{(ans)}P^{(sup)}, \  R^{(joint)} = R^{(ans)}R^{(sup)},
\end{equation}
\begin{equation}
	\textrm{Joint F}_1 = \frac{2 P^{(joint)} R^{(joint)}}{P^{(joint)} \times R^{(joint)}},
\end{equation}
where $P^{(ans)}$ and $P^{(sup)}$ are the precisions of the answer span and the support facts for each, and $R^{(ans)}$ and $R^{(sup)}$ are the recalls of the answer span and the support facts for each. This evaluation metric forces machines to find not only the correct answer span but also the correct support facts. 

Wikireading~\citep{wikireading2016} is the largest reading comprehension dataset in the datasets in this section that consists of 19M pairs of a question and answer. The dataset is constructed from Wikipedia and Wikidata. Wikipedia is a free online encyclopedia hosted by the Wikimedia Foundation that consists of more than 6 million articles\footnote{https://en.wikipedia.org/wiki/English\_Wikipedia (last accessed June 2020)}. Wikidata is a collaboratively edited knowledge base hosted by the Wikimedia Foundation that consists of sets of tuples, i.e., (subject entity, relation type, argument entity). There are more than 7,000 relation types, including ``instance\_of'' and ``location'', and most entities in Wikidata and entries in Wikipedia are linked for each other. In the dataset, each question is a pair of the subject entity and relation type in a tuple, and then the answer is the argument entity in the tuple. The corresponding passage for the question is a Wikipedia article whose title is the subject entity. The answer type is free-form answer, and a machine is expected to predict the name of the argument entity. Again, the evaluation metrics are EM and F1 as following SQuAD. The dataset is pretty biased, and the top 20 relation types cover 75\% of the dataset so that the dataset might not require general reading comprehension skills. 

WikiHop~\citep{welbl-etal-2018-constructing} is a reading comprehension dataset aiming for multihop reading comprehension. Multihop reading comprehension is a reading comprehension task where the question cannot be solved by any single sentence in the given passage, but it can be solved by information written in multiple sentences. We call the reading comprehension skill at getting together the information written in multiple sentences as the multihop inference. Similar to Wikireading, each question of Wikihop consists of a subject entity and relation type, but the passage is a set of paragraphs from multiple Wikipedia articles to encourage the multihop inference. Additionally, each question provides candidate answers so that it is multiple-choice question answering task. We describe the detail of the dataset in Section \ref{sec:wikihop}. 

\begin{table*}[t!]
	\centering
	\fbox{\begin{minipage}[t]{\textwidth}

\textbf{Passage:} The Hanging Gardens, in Mumbai, also known as Pherozeshah Mehta Gardens, are terraced gardens … They provide sunset views over the Arabian Sea...

Mumbai (also known as Bombay, the official name until 1995) is the
capital city of the Indian state of Maharashtra. It is the most
populous city in India...

The Arabian Sea is a region of the northern Indian Ocean bounded
on the north by Pakistan and Iran, on the west by northeastern
Somalia and the Arabian Peninsula, and on the east by India ...

\textbf{Question:} (Hanging gardens of Mumbai, country, \#BLANK\#)

\textbf{Candidate answers:} a)Iran, b)India, c)Pakistan, d)Somalia, ...

\textbf{Answer:} (b)India

    \end{minipage}}
    \caption{A sample question from Wikihop dataset. }
    \label{table:wikihop.sample}
\end{table*}

\begin{table}[t]
\centering
\begin{tabular}{l||lll}
  \toprule
Dataset & Answer type & Text resource & Data size  \\ \midrule
Deep Read dataset   & Sentence selection & 3rd to 6th grade material & 60 {\scriptsize stories} $\times$ 5 {\scriptsize questions}\\ \hline
MCTest              & Multiple choice & Fictional story & 2640 questions \\
CNN/Daily Mail      & Multiple choice & News article & 1.4M questions\\
Children Book Test  & Multiple choice & Children Book & 687K questions \\
WDW       & Multiple choice & News article & 206K questions\\
WikiHop             & Multiple choice & Wikipedia and Wikidata & 51K questions\\ \hline
SQuAD               & Span prediction & Wikipedia & 100K questions\\
HotpotQA            & Span prediction & Wikipedia & 16K-91K questions\\ \hline
TriviaQA            & Free-form answer & Wikipedia and Web-page & 96K questions\\
NarrativeQA         & Free-form answer & Book and movie script & 47K questions\\
Wikireading         & Free-form answer & Wikipedia & 13M questions\\
\bottomrule
\end{tabular}
\caption{Notable reading comprehension datasets since the 1990s. }
\label{table:wdw.related.size}
\end{table}


\section{Dataset construction}

\label{sec:details}
We now describe the construction of our WDW in more detail. To generate a question, we first generate the question by selecting a random article --- the ``question article'' --- from the Gigaword corpus and taking the first sentence of that article --- the ``question sentence'' --- as the source of the cloze question. The hope is that the first sentence of an article contains prominent people and events which are likely to be discussed in other independent articles. To convert the question sentence to a cloze question, we first extract named entities using the Stanford NER system~\citep{manning-EtAl:2014:P14-5} and parse the sentence using the Stanford PCFG parser~\citep{klein2003}.

The person named entities are candidates for deletion to create a cloze problem. For each person named entity, we then identify a noun phrase in the automatic parse that is headed by that person. For example, if the question sentence is ``President Obama met yesterday with Apple Founder Steve Jobs'' we identify the two person noun phrases ``President Obama'' and ``Apple Founder Steve Jobs''.  When a person named entity is selected for deletion,  the entire noun phrase is deleted.  For example, when deleting the second named entity, we get ``President Obama met yesterday with ***'' rather than ``President Obama met yesterday with Apple founder ***''. This increases the difficulty of the problems because systems cannot rely on descriptors and other local contextual cues. About 700,000 question sentences are generated from Gigaword articles (8\% of the total number of articles).

Once a cloze question has been formed, we select an appropriate article as a passage.  The article should be independent of the question article but should discuss the people and events mentioned in the question sentence. To find a passage, we search the Gigaword dataset using the Apache Lucene information retrieval system~\citep{McCandless2010LuceneEdition}, using the question sentence as the query. The named entity to be deleted is included in the query and required to be included in the returned article.  We also restrict the search to articles published within two weeks of the date of the question article. Articles containing sentences too similar to the question in word overlap and phrase matching near the blanked phrase are removed. We select the best matching article satisfying our constraints.  If no such article can be found, we abort the process and move on to a new question.


Given a question and a passage, we next form the list of choices.  We collect all person named entities in the passage except unblanked person named entities in the question. Person named entities that are subsets of another longer named entity are eliminated from the choice list. For example, the choice ``Obama'' would be eliminated if the list also contains ``Barack Obama''. We also discard ambiguous cases where a part of a blanked NE appears in multiple choices in the list, e.g., if a passage has ``Bill Clinton'' and ``Hillary Clinton'' and the blanked phrase is ``Clinton'' then we discard it. We found this simple coreference rule to work well in practice since news articles usually employ full names for initial mentions of persons. If the resulting choice list contains fewer than two or more than five choices, the process is aborted and we move on to a new question.\footnote{The maximum of five helps to avoid sports articles containing structured lists of results.}


After forming an initial set of problems, we then remove ``duplicated'' problems.  Duplication arises because Gigaword contains many copies of the same article or articles where one is clearly an edited version of another.  Our duplication-removal process ensures that no two problems have very similar questions. Here, similarity is defined as the ratio of the size of the bag of words intersection to the size of the smaller bag.


Then we remove some problems in order to focus our dataset on the most interesting problems. We decided to remove questions that can be solved by a syntactic matching algorithm, counting algorithm, or simple heuristic algorithm because we found machine learning systems easily learned these techniques from these questions; thus, they were not appropriate to teach and test deeper reading comprehension skills of these machine learning systems. 
We used the following two syntactic matching algorithms, a counting algorithm, and a heuristic algorithm as baselines to find such questions. We remove these questions to suppress their performance.
\begin{itemizesquish}
    \item First person in passage: Select the person that appears first in the passage.
    \item Most frequent person: Select the most frequent person in the passage.
    \item $n$-gram: Select the most likely answer to fill the blank under a 5-gram language model trained on Gigaword minus articles which are too similar to one of the questions in word overlap and phrase matching.
    \item Unigram: Select the most frequent last name using the unigram counts from the 5-gram model.
\end{itemizesquish}
To minimize the number of questions removed we solve an optimization problem defined by limiting the performance of each baseline to a specified target value while removing as few problems as possible, i.e.,
\begin{equation}
    \max_{\alpha(C)} \sum_{C \in \{0,1\}^{|b|}} \alpha(C) | T(C) |,
\end{equation}
subject to
\begin{eqnarray}
    &\forall i& \ \sum_{ C : C_i = 1} \frac{\alpha(C) | T(C)| }{N} \le k, \nonumber \\
    &N& = \sum_{C \in \{0,1\}^{|b|}} \alpha(C) | T(C) |,
\end{eqnarray}
where $T(C)$ is the subset of the questions solved by the subset $C$ of the suppressed baselines, $\alpha(C)$ is a keeping rate for question set $T(C)$, $C_i=1$ indicates the $i$-th baseline is in the subset, $|b|$ is the number of baselines, $N$ is a total number of questions, and $k$ is the upper bound for the baselines after suppression. We choose $k$ to yield random performance for the baselines. The performance of the baselines before and after suppression is shown in Table~\ref{table:beforeKnockOff}.
The suppression removed 49.9\% of the questions. 

Table~\ref{table:statistics} shows statistics of our dataset after suppression.  We split the final dataset into train, validation, and test by taking the validation and test to be a random split of the most recent 20,000 problems as measured by question article date.  In this way there is very little overlap in semantic subject matter between the training set and either validation or test. We also provide a larger ``relaxed'' training set formed by applying less baseline suppression (a larger value of $k$ in the optimization).  The relaxed training set then has a slightly different distribution from the train, validation, and test sets which are all fully suppressed.

\begin{table}[t]
\centering
\begin{tabular}{@{}l|ll@{}}
  \toprule
        & \multicolumn{2}{c}{Accuracy}                          \\
\multicolumn{1}{c|}{Baseline} & \multicolumn{1}{c}{Before} & \multicolumn{1}{c}{After} \\ \midrule
First person in passage                &  0.60   &    0.32                      \\
Most frequent person              &  0.61   &    0.33                       \\
$n$-gram                         &  0.53   &    0.33                       \\
Unigram                       &  0.43   &    0.32                      \\
Random$^\ast$ & 0.32 & 0.32 \\
\bottomrule
\end{tabular}
\caption{Performance of suppressed baselines. $^\ast$Random performance is computed as a deterministic function of the number of times each choice set size  appears. Many questions have only two choices and there are about three choices on average.
}
\label{table:beforeKnockOff}
\end{table}

\begin{table}[t]
\centering
\begin{tabular}{@{}lrrrr@{}} \toprule
 & relaxed train & train & valid & test   \\ \midrule
\# questions         & 185,978     & 127,786   & 10,000  & 10,000  \\
avg. \# choices     & 3.5   & 3.5      & 3.4    & 3.4    \\
avg. \# tokens      & 378 & 365    & 325  & 326  \\
vocab. size         & 347,406 & \multicolumn{3}{c}{308,602} \\
\bottomrule
\end{tabular}
\caption{Dataset statistics.}
\label{table:statistics}
\end{table}

\section{Performance Benchmarks}

\label{sec:baselines}
We report the performance of following several systems to characterize our dataset:
\begin{itemize}
    \item Word overlap: Select the choice $c$ inserted to the question $q$ which is the most similar to any sentence $s$ in the passage, i.e., ${\rm CosSim}({\rm bag}(c+q), {\rm bag}(s))$.
    \item Sliding window and Distance baselines (and their combination) from \citet{Richardson2013}.
    \item Semantic features: NLP feature based system from \citet{haiwang2015}.
    \item Attentive Reader: LSTM with attention mechanism \citep{Hermann2015}.
    \item Stanford Reader: An attentive reader modified with a bilinear term \citep{chen2016}.
    \item Attention Sum Reader: GRU with a point-attention mechanism \citep{Kadlec2016}.
    \item Gated-Attention Reader: Attention Sum Reader with gated layers \citep{Dhingra2016}.
\end{itemize}

\noindent Table~\ref{table:baseline} shows the performance of each system on the test data. For the Attention and Stanford Readers, we anonymized the WDW data by replacing named entities with entity IDs as in the CNN/Daily Mail dataset. 

We see consistent reductions in accuracy when moving from CNN to our dataset. The Attentive and Stanford Reader drop by up to 10\% and the Attention Sum and Gated-Attention readers drop by up to 17\%. The ranking of the systems also changes. In contrast to the Attentive/Stanford readers, the Attention Sum/Gated-Attention readers explicitly leverage the frequency of the answer in the passage, a heuristic which appears beneficial for the CNN/Daily Mail tasks. It seems that our suppression of the most-frequent-person baseline more strongly affects the performance of these latter systems.

\begin{table}[ht]
\centering
\begin{tabular}{@{}lrr@{}}
\toprule
System & WDW & CNN\\ \midrule
Word overlap & 0.47 & -- \\
Sliding window & 0.48 & -- \\
Distance & 0.46 & -- \\
Sliding window + Distance  &  0.51 & -- \\
Semantic features & 0.52 & -- \\ \hdashline 
Attentive Reader & 0.53 & $0.63^{I}$ \\
Attentive Reader (relaxed train) & 0.55 &  \\ 
Stanford Reader & 0.64 & $0.73^{I\hspace{-.1em}I}$ \\
Stanford Reader (relaxed train) & 0.65 &  \\
Attention Sum Reader & 0.57 & $0.70^{I\hspace{-.1em}I\hspace{-.1em}I}$ \\ 
Attention Sum Reader (relaxed train) & 0.59  &  \\
Gated-Attention Reader & 0.57 & $0.74^{I\hspace{-.1em}V}$ \\
Gated-Attention Reader (relaxed train) & 0.60 &  \\ \hline 
Human Performance & $84/100$ & $0.75+^{I\hspace{-.1em}I}$ \\ \bottomrule
\end{tabular}
\caption{System performance on test set. Human performance was computed by two annotators on a sample of 100 questions. Result marked ${I}$ is from \citet{Hermann2015}, results marked ${I\hspace{-.1em}I}$ are from \citet{chen2016}, result marked $I\hspace{-.1em}I\hspace{-.1em}I$ is from \citet{Kadlec2016}, and result marked $I\hspace{-.1em}V$ is from \citet{Dhingra2016}. }
\label{table:baseline}
\end{table}

\section{Conclusion}

We presented a large-scale person-centered cloze dataset. The dataset is not anonymized, and each passage is a raw text which is not only natural but also easier to be pre-processed by syntactic and semantic parsers. In the dataset construction, we used baseline suppression, where we selected undesired questions by multiple baseline systems and randomly removed some of them. This approach can flexibly design the difficulty and quality of a dataset by replacing baseline systems that select undesired questions. 
As a result, we obtained about 200M questions and achieved the higher human performance and the lower machine performance, and then the larger performance gap between them. This result indicates that the dataset requires deeper reading comprehension skills that these machines do not have. This dataset is different in a variety of ways from existing large-scale cloze datasets and provides a significant extension to the training and test data for machine comprehension.

\chapter{Analysis of a neural structure in entity-centered reading comprehension}  \label{sec:work2}


As we discussed in Section \ref{sec:wdw:related}, several large scale cloze-style reading comprehension datasets~\citep{Hermann2015, Hill2016TheRepresentations, Onishi2016} have been introduced, and the large sizes of them enable the application of deep learning. Despite the significant performance of the deep learning models, the prediction structure of these models is poorly understood.

In this chapter, we present empirical evidence for the emergence of predication structure in a certain class of deep learning models for reading comprehension (neural readers); ``Aggregation'' and ``Explicit reference'' readers. Both readers work on the CNN/Daily Mail dataset, a dataset with anonymized entities. This work was published as the best paper in 2nd Workshop on Representation Learning for NLP~\citep{wang-etal-2017-emergent}.

Before we explain the neural readers, we review the CNN/Daily Mail dataset where entities are anonymized. This dataset consists of anonymized passages and questions where named entities are replaced by anonymous entity identifiers such as ``entity37''. For example, the passage might contain ``entity52 gave entity24 a rousing applause'', and the question might be ``$X$ received a rounding applause from entity52'', then the answer is the most appropriate entity identifier in the passage to fill $X$. The same entity identifiers are used over all the problems, and a different identifier is assigned to an entity every time the passage and question are read. Thus, the entity identifiers are presumably just pointers to semantics-free tokens and do not have any semantic meaning. We will write entity identifiers as logical constant symbols such as $c$ rather than strings such as ``entity37''.

``Aggregation'' readers, including Memory Networks~\cite{Weston2015, Sukhbaatar2015End-To-EndNetworks}, the Attentive Reader~\citep{Hermann2015},  and the Stanford Reader~\citep{chen2016}, use bidirectional LSTMs or GRUs to construct a contextual embedding $h_t$ of each position $t$ in the passage and also an embedding $h_q$ of the question $q$.  They then select an answer $c$ using a criterion similar to
\begin{equation}
  \label{eqn1}
  \argmax_c  \; \sum_t\;   \left<h_t,h_{q}\right>\;\;\; \left<h_t,e(c)\right>,
\end{equation}
where $e(c)$ is the vector embedding of the constant symbol (entity identifier) $c$.  In practice the inner-product $\langle h_t,h_{q}\rangle$ is normalized over $t$ using a softmax to yield attention weights $\alpha_t$ over $t$ and Equation (\ref{eqn1}) becomes
\begin{equation}
  \label{eqn2}
  \argmax_c  \; \left<e(c),\;\sum_t\; \alpha_t h_t\right>.
\end{equation}
Here $\sum_t\;\alpha_t h_t$ can be viewed as a vector representation of the passage.

We argue that for aggregation readers, roughly defined by Equation (\ref{eqn2}). Letting the $t$-th hidden state of the passage be $h_t$, the state is a contextual embedding of the $t$-th token and can be viewed as a vector concatenation $h_t = [s(\Phi_t), s(c_t)]$ where $\Phi_t$ is a property (or statement or predicate) being stated of a particular constant symbol $c_t$.  Here $s(\Phi_t)$ and $s(c_t)$ are unknown emergent embeddings of $\Phi_t$ and $c_t$ respectively. A logician might write this as $h_t = \Phi_t[c_t]$. Furthermore, the question can be interpreted as having the form $\Psi[x]$ where the problem is to find a constant symbol $c$ such that the passage implies $\Psi[c]$. Assuming $h_t = [s(\Phi_t),s(c_t)]$, $h_{q} = [s(\Psi),0]$, and $e(c) = [0,s(c)]$, we can rewrite Equation (\ref{eqn1}) as
\begin{equation}
  \label{eqn3}
  \argmax_c   \; \sum_t \;    \left<s(\Phi_t),s(\Psi) \right>\;\;\;  \left<s(c_t),s(c)\right>.
\end{equation}
The first inner product in Equation (\ref{eqn3}) is interpreted as measuring the extent to which $\Phi_t[x]$ implies $\Psi[x]$ for any $x$.
The second inner product is interpreted as restricting $t$ to positions talking about the constant symbol $c$.
Note that the posited decomposition of $h_t$ is not explicit in Equation (\ref{eqn2}) but instead must emerge during training.  We present empirical evidence that this structure does emerge.  The empirical evidence is somewhat tricky as the direct sum structure that divides $h_t$ into its two parts need not be axis aligned and therefore need not literally correspond to vector concatenation.

``Explicit reference readers'', including the Attention Sum Reader~\cite{Kadlec2016}, the Gated-Attention Reader~\cite{Dhingra2016}, and the Attention-over-Attention Reader~\cite{CuiAttention-over-AttentionComprehension}, avoid Equation (\ref{eqn2}) and instead use
\begin{equation}
  \label{eqn4}
  \argmax_c \sum_{t \in R(c)} \alpha_t,
\end{equation}
where $R(c)$ is the subset of the positions where the constant symbol (entity identifier) $c$ occurs.  Note that if we identify $\alpha_t$ with $\left<s(\Phi_t),s(\Psi)\right>$ and assume that $\left<s(c),s(c_t)\right>$ is either 0 or 1 depending on whether $c = c_t$, then Equations (\ref{eqn3}) and (\ref{eqn4}) agree. In explicit reference readers, the hidden state $h_t$ need not carry a pointer to $c_t$ as the restriction on $t$ is independent of learned representations. 

In this research, we have only considered anonymized datasets that require the handling of semantics-free constant symbols.  However, even for non-anonymized datasets such as WDW, it is helpful to add features which indicate which positions in the passage are referring to which candidate answers. This indicates, not surprisingly, that reference is important in question answering.  The fact that explicit reference features are needed in aggregation readers on non-anonymized data indicates that reference is not being solved by the aggregation readers.  However, as reference seems to be important for cloze-style question answering, these problems may ultimately provide training data from which reference resolution can be learned.

\section{Related work}

Here we classify readers into aggregation readers and explicit reference readers. Aggregation readers appeared first in the literature, including Memory Networks~\citep{Weston2015,Sukhbaatar2015End-To-EndNetworks}, the Attentive Reader~\cite{Hermann2015}, and the Stanford Reader~\citep{chen2016}. Then, Explicit reference readers, including the Attention Sum Reader~\citep{Kadlec2016}, the Gated-Attention Reader~\citep{Dhingra2016}, and the Attention-over-Attention Reader~\citep{CuiAttention-over-AttentionComprehension}, were proposed. In the following sections, we define aggregation readers more specifically by Equations (\ref{eqn:r}) and (\ref{eqn:testloss}) and then explicit reference readers by Equation (\ref{eqn:testloss-AS}).  We first present the Stanford Reader as a paradigmatic aggregation reader and the Attention Sum Reader as a paradigmatic explicit reference reader.

\subsubsection{Aggregation Readers} \label{sec:aggregation_reader}

\hspace{1.0em}\\ \noindent
\textbf{Stanford Reader.} The Stanford Reader~\citep{chen2016} computes a bidirectional LSTM~\citep{Hochreiter1997} representation of both the passage and the question.
\begin{eqnarray}
  \label{eqn:h}
  h & = & \text{biLSTM}(e(p)). \\
  \label{eqn:q}
  h_{q} & = & [\text{fLSTM}(e(q))_{|q|},\text{bLSTM}(e(q))_1].
\end{eqnarray}
In Equations (\ref{eqn:h}) and (\ref{eqn:q}), $e(p)$ is the sequence of word embeddings $e(w_i)$ for $w_i \in p$ and similarly for $e(q)$.  The expression $\text{biLSTM}(s)$ denotes the sequence of hidden state vectors resulting from running a bidirectional LSTM on the vector sequence $s$.  We write $\text{biLSTM}(s)_i$ for the $i$-th vector in this sequence.  Similarly $\text{fLSTM}(s)$ and $\text{bLSTM}(s)$ denote the sequence of vectors resulting from running a forward LSTM and a backward LSTM respectively and $[\cdot,\cdot]$ denotes vector concatenation. The Stanford Reader, and various other readers, then compute a bilinear attention over the passage which is used to construct a single weighted vector representation of the passage. 
\begin{eqnarray}
  \label{eqn:attention}
  \alpha_{t}  =  \softmax_{t} \;h_{t}^\top W_\alpha \;h_{q}, \quad
  \label{eqn:r}
  o  =  \sum_t \alpha_t h_t.
\end{eqnarray}
Finally, they compute a probability distribution $P$ over the answers:
\begin{eqnarray}
  \label{eqn:trainloss}
  P(\cdot|d,q,{\cal A}) & = & \softmax_{a \in {\cal A}} \; e_o(a)^{\top}o, \\
  \label{eqn:testloss}
  \hat{a} & = & \argmax_{a \in \mathcal{A}}\; e_o(a)^\top o .
\end{eqnarray}
Here $e_o(a)$ is the ``output embedding'' of the answer $a$.  On the CNN/Daily Mail dataset the Stanford Reader learns an output embedding for each of the roughly 550 entity identifiers used in the dataset. For datasets in which the answer might be any word in ${\cal V}$, output embeddings must be trained for the entire vocabulary. The reader is trained with log-loss $-\log P(a|p,q,{\cal A})$ where $a$ is the correct answer. At test time the reader is scored on the percentage of problems where $\hat{a} = a$.

\hspace{1.0em}\\ \noindent
{\bf Memory Networks.} Memory Networks~\cite{Weston2015,Sukhbaatar2015End-To-EndNetworks} use Equations (\ref{eqn:r}) and (\ref{eqn:testloss}) but have more elaborate methods of
constructing ``memory vectors'' $h_t$ not involving LSTMs. Memory networks use Equations (\ref{eqn:r}) and (\ref{eqn:testloss}) but replace Equation (\ref{eqn:trainloss}) with
\begin{equation}
  \label{eqn:trainloss-wide}
  P(\cdot|p,q,{\cal A}) = P(\cdot|p,q) = \softmax_{w \in {\cal V}} e_o(w)^\top o.
\end{equation}
Note that Equation (\ref{eqn:trainloss-wide}) trains output vectors over the whole vocabulary rather than just those items occurring in the choice set ${\cal A}$. This is empirically significant in non-anonymized datasets such as CBT and WDW where choices at test time may never have occurred as choices in the training data.

\hspace{1.0em}\\ \noindent
{\bf Attentive Reader.} The Stanford Reader was derived from the Attentive Reader~\cite{Hermann2015}. The Attentive Reader uses $\alpha_t = \softmax_t \text{MLP}([h_t,h_{q}])$ instead of Equation (\ref{eqn:attention}). Here $\text{MLP}(x)$ is the output of a multi layer perceptron given input $x$. Also, the answer distribution in the Attentive Reader is defined over the full vocabulary rather than just the candidate answer set ${\cal A}$:
\begin{equation}
  \label{eqn:trainloss-wide2}
P(\cdot|p,q,{\cal A}) = \softmax_{w \in {\cal V}}\: e_o(w)^\top \text{MLP}([o,h_{q}]).
\end{equation}
Equation (\ref{eqn:trainloss-wide2}) is similar to Equation (\ref{eqn:trainloss-wide}) in that it leads to the training of output vectors for the full vocabulary rather than just those items appearing in choice sets in the training data.  As in memory networks, this leads to improved performance on non-anonymized datasets.

\subsubsection{Explicit Reference Readers}

\hspace{1em} \\ \noindent
{\bf Attention Sum Reader.} In the Attention Sum Reader~\cite{Kadlec2016}, $h$ and $q$ are computed with Equations (\ref{eqn:h}) and (\ref{eqn:q}) as in the Stanford Reader but using GRUs rather than LSTMs. The attention $\alpha_t$ is computed similarly to Equation (\ref{eqn:attention}) but using a simple inner product $\alpha_{t} = \softmax_{t} \;h_{t}^\top h_{q}$ rather than a trained bilinear form.  Most significantly, however, Equations (\ref{eqn:trainloss}) and (\ref{eqn:testloss}) are replaced by
the following where $t \in R(a,p)$ indicates that a reference to the candidate answer $a$ occurs at the position $t$ in $p$.
\begin{eqnarray}
  \label{eqn:trainloss-AS}
  P(a|p,q,{\cal A}) & = & \sum_{t \in R(a,p)} \;\alpha_t . \\
  \label{eqn:testloss-AS}
  \hat{a} & = & \argmax_a \sum_{t \in R(a,p)} \;\alpha_t . 
\end{eqnarray}
Here we think of $R(a,p)$ as the set of references to $a$ in the passage $p$.  It is important to note that Equation (\ref{eqn:trainloss-AS}) is an equality and that $P(a|p,q,{\cal A})$ is not normalized to the members of $R(a,p)$.  When training with the log-loss objective this drives the attention $\alpha_t$ to be normalized --- to have support only on the positions $t$ with $t \in R(a,p)$ for some $a$.

\hspace{1em} \\ \noindent
\textbf{Gated-Attention Reader.} The Gated-Attention Reader~\cite{Dhingra2016} involves a $K$-layer biGRU architecture defined by the following equations.
\begin{eqnarray}
  h_{q}^{\ell} & = & [\text{fGRU}(e(q))_{|q|},\text{bGRU}(e(q))_{1}] , \;\;1 \leq \ell \leq K .\\
  h^1 & = & \text{biGRU}(e(p)) . \\
  h^{\ell} & = & \text{biGRU}(h^{\ell-1} \odot h_{q}^{\ell-1}) ,\;\;2 \leq \ell \leq K .
\end{eqnarray}
Here the question embeddings $h_{q}^\ell$ for different values of $\ell$ are computed with different GRU model parameters.  Here $h \odot h_{q}$ abbreviates the sequence $h_1\odot h_{q}$, $h_2 \odot h_{q}$, $\ldots$ $h_{|p|}\odot h_{q}$. Note that for $K=1$ we have only $h_{q}^1$ and $h^1$ as in the Attention Sum Reader. An attention is then computed over the final layer $h^K$ with $\alpha_{t} = \softmax_{t} \;(h^K_{t})^\top \;h_{q}^{K}$ in the Attention Sum Reader. This reader uses Equations (\ref{eqn:trainloss-AS}) and (\ref{eqn:testloss-AS}).

\hspace{1em} \\ \noindent
\textbf{Attention-over-Attention Reader.} The Attention-over-Attention Reader~\cite{CuiAttention-over-AttentionComprehension} uses a more elaborate method to compute the attention $\alpha_t$. We will use $t$ to range over positions in the passage and $j$ to range over positions in the question.  The model is then defined by the following equations.
$$\begin{array}{cc}
  h = \text{biGRU}(e(p)) , & h_{q} = \text{biGRU}(e(q)) . \\
  \\
  \alpha_{t,j} = \softmax_t h_t^\top h_{q,j} ,&\beta_{t,j} = \softmax_j h_t^\top h_{q,j} .\\
  \\
\beta_j = \frac{1}{|p|} \sum_t \beta_{t,j} ,& \alpha_t = \sum_j \beta_j \alpha_{t,j} .
\end{array}$$
Note that the final equation defining $\alpha_t$ can be interpreted as applying the attention $\beta_j$ to the attentions $\alpha_{t,j}$. This reader uses Equations (\ref{eqn:trainloss-AS}) and (\ref{eqn:testloss-AS}).

\section{Emergent Predication Structure}
\label{analysis}
In this section, we claim an emergent predication structure in the hidden vectors $h_t$ that explains the high performance of aggregation readers. Intuitively we think of the hidden state vector $h_t$ as a concatenation $[s(\Phi_t),s(a_t)]$ where $\Phi_t$ is a property being asserted of entity $a_t$ at the position $t$ in the passage.  Here $s(\Phi_t)$ and $s(a_t)$ are emergent embeddings of the property and entity respectively, we also think of the vector representation $q$ of the question as having the form $[s(\Psi),0]$ and the vector embedding $e_o(a)$ of an entity as having the form $[0,s(a)]$. Remember that the vector embeddings have no semantics as discussed, and they are considered as pointers or semantics-free constant symbols.

Formally, the decomposition of $h_t$ into this predication structure is not necessarily axis aligned.  Rather than posit an axis-aligned concatenation, we posit that the hidden vector space $H$ is a possibly non-aligned direct sum
\begin{equation}  \label{eqn:predict3}
  H = S \oplus E .
\end{equation}
where $S$ is a subspace of ``statement vectors'' and $E$ is an orthogonal subspace of ``entity pointers''. Each hidden state vector $h \in H$ then has a unique decomposition as $h = \Psi + e$ for $\Psi \in S$ and $e \in E$.  This is equivalent to saying that the hidden vector space $H$ is some rotation of a concatenation of the vector spaces $S$ and $E$.  In this non-axis aligned model, we assume emergent embeddings $s(\Phi)$ and $s(a)$ with $s(\Phi) \in S$ and $s(a) \in E$.  We also assume that the latent spaces are learned in such a way that explicit entity output embeddings satisfy $e_o(a) \in E$.

This predication structure explains that a question asks for a value of $x$ such that a statement $\Psi[x]$ is implied by the passage. For a question $\Psi$ we might even suggest the following vectorial interpretation of entailment. 
\begin{equation}
    \Phi[x]\;\mbox{implies} \;\Psi[x] \;\;\; \mbox{iff}\;\;\;  \Phi^\top \Psi \geq ||\Psi||_1.
\end{equation}
This interpretation is exactly correct if some of the dimensions of the vector space correspond to predicates, $\Psi$ is a 0-1 vector representing a conjunction predicates, and $\Phi$ is also 0-1 on these dimensions indicating whether a predicate is implied by the context.

We now present empirical evidence for this emergent structure. The empirical evidence supports two corollaries that are derived from the structure. 

\hspace{1em} \\ \noindent
\textbf{Corollary A:} For some fixed positive constant $c$, 
\begin{equation} \label{eqn:predict1}
  e_o(a)^\top h_t = \left\{\begin{array}{ll} c & \mbox{if}\;t \in R(a,p) \\
      0 & \mbox{otherwise} .
    \end{array}\right.
\end{equation}
We note that if $e_o(a)^\top s(a)$ was different for each candidate answer $a$ then answers would be biased toward constant symbols where this product was larger. This contradicts the anonymization of entity identifiers, and then all constant symbols must be equivalent. 
It is also worth mentioning that Corollary A makes Equations (\ref{eqn:testloss}) and (\ref{eqn:testloss-AS}) agree as follows: 
\begin{eqnarray}
\argmax_a \;e_o(a)^\top o  &=&  \argmax_a\;e_o(a)^\top \sum_t\; \alpha_t h_t \\
        &=&  \argmax_a\;\sum_t\; \alpha_t\; e_o(a)^\top h_t \\
        &=& \argmax_a \sum_{t \in R(a,p)} \alpha_t.
\end{eqnarray}
Thus, the aggregation readers and the explicit reference readers are using essentially the same answer selection criterion.

The first three rows of Table \ref{table: statistics} is empirical evidence for Corollary A. The first row empirically measures the constant $c$ in Equation (\ref{eqn:predict1}) by measuring $e_o(a)^\top h_t$ for those cases where $t \in R(a,p)$.  The second row measures ``0'' in Equation (\ref{eqn:predict1}) by measuring $e_o(a)^\top h_t$ in those cases where $t \not \in R(a,p)$.  The third row shows that this inner product falls off significantly just one word before or after the position of the answer word. 

\begin{figure}[t]
\centering
\includegraphics[width=.75\linewidth]{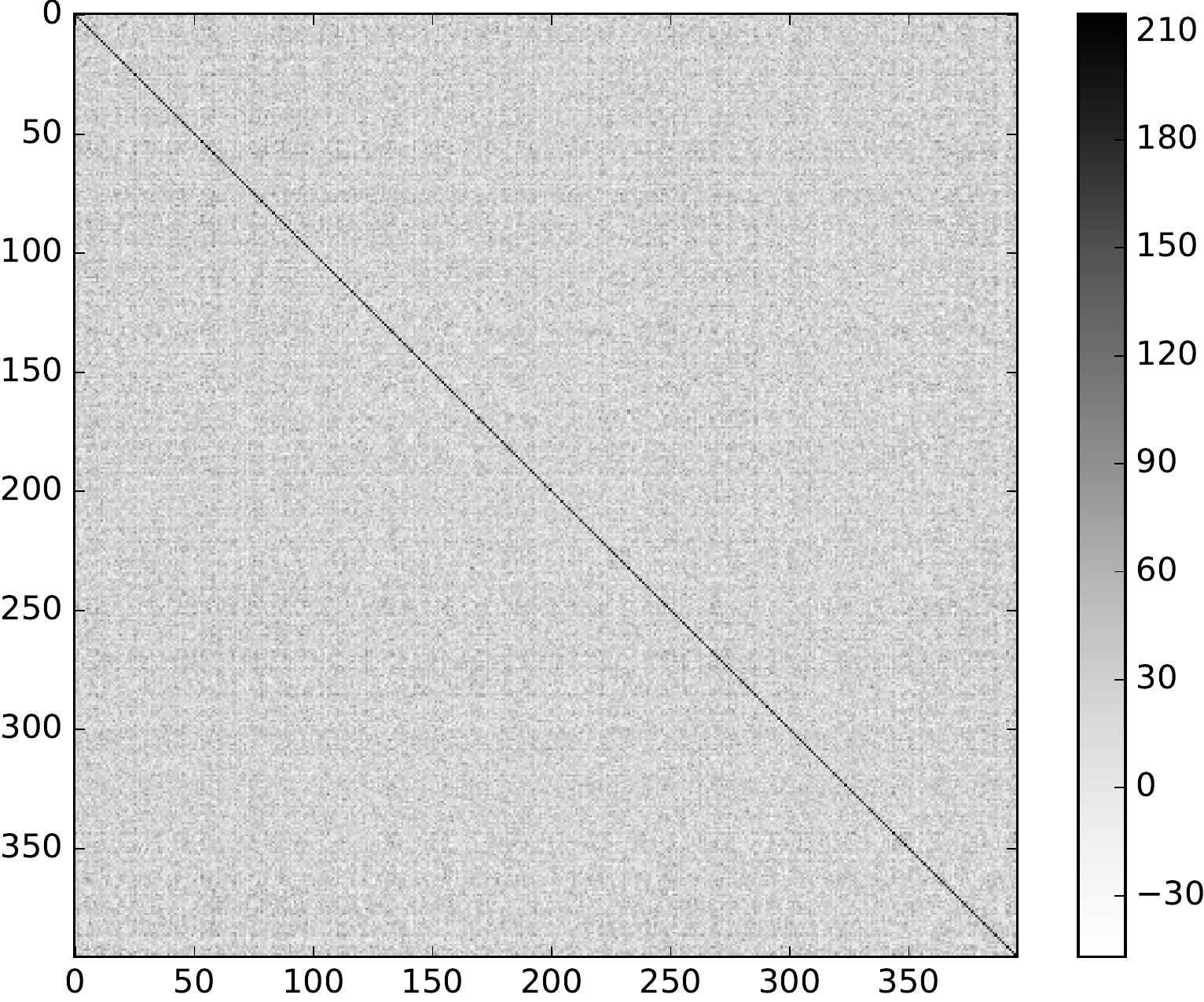}
\caption{Plot of $e_o(a_i)^\top e_o(a_j)$ from the Stanford Reader trained on the CNN dataset, where rows range over $i$ values and columns range over $j$ values. Off-diagonal values have mean 25.6 and variance 17.2 while diagonal values have mean 169 and variance 17.3.}
\label{fig:diag}
\end{figure}

Figure \ref{fig:diag} shows that the output vectors $e_o(a)$ for different entity identifiers $a$ are nearly orthogonal. The orthogonality of the output vectors is required by Equation (\ref{eqn:predict1}) provided that each output vector $e_o(a)$ is in the span of the hidden state vectors $h_{t,p}$ for which $t \in R(a,p)$.  Intuitively, the mean of all vectors $h_{t,p}$ with $t \in R(a,p)$ should be approximately equal to $e_o(a)$. Empirically this will only be approximately true. 

Theoretically, Corollary A would suggest that the vector embedding of the constant symbols should have the number of dimensions at least as large as the number of distinct constants. However, it is sufficient that $e_o(a)^\top s(a')$ is small for $a \not = a'$ to make the neural readers work in practice, and this also allows the vector embeddings of the constants to have dimension much smaller than the number of constants. We have experimented with two-sparse constant symbol embeddings where the number of embedding vectors in dimension $d$ is $2d(d-1)$ ($d$ choose 2 times the four ways of setting the signs of the non-zero coordinates).  Although we do not report results here, these designed and untrained constant embeddings worked reasonably well.

\hspace{1em} \\ \noindent
\textbf{Corollary B:} 
\begin{equation} \label{eqn:predict2}
  h_{q}^\top(h_i + e_o(a)) = h_{q}^\top h_i.
\end{equation}
This equation is equivalent to $h_{q}^\top e_o(a) = 0$.  Experimentally, however, we cannot expect $h_{q}^\top e_o(a)$ to be exactly zero and Equation (\ref{eqn:predict2}) seems to provides a more experimentally meaningful test. 

The fourth and fifth rows of Table \ref{table: statistics} is an empirical evidence for Corollary B. The fourth row measures the cosine of the angle between the question vector $h_{q}$ and the hidden state $h_t$ averaged over passage positions $t$ at which some entity identifier occurs. The fifth row measures the cosine of the angle between $h_{q}$ and $e_o(a)$ averaged over the entity identifiers $a$.

\begin{table}[t]
\begin{center}
\resizebox{\textwidth}{!}{\begin{tabular}{@{}l|rrr|rrr} \toprule
   & & {CNN Dev} & & & {CNN Test} & \\
   & samples&mean&variance&samples&mean&variance\\ \hline
   $e_o(a)^\top h_t, \quad t \in R(a, p)$     & 222,001&10.66 &2.26 & 164,746&10.70 &2.45 \\
   $e_o(a)^\top h_t, \quad t \notin R(a, p)$ & 93,072,682&-0.57 &1.59 & 68,451,660&  -0.58 & 1.65 \\
   $e_o(a)^\top h_{t\pm 1}, \quad t \in R(a, p)$     & 443,878 &2.32 &1.79 & 329,366& 2.25 &1.84 \\ \hline
   $\mathrm{Cosine}(h_{q}, h_t),  \quad \exists a\;t \in R(a, p)$ & 222,001& 0.22 & 0.11 & 164,746& 0.22&0.12 \\
   $\mathrm{Cosine}(h_{q}, e_o(a)),\quad \forall a $ & 103,909& -0.03 & 0.04 & 78,411& -0.03& 0.04\\ \bottomrule
   
          
 \end{tabular}}
\end{center}
\caption{Statistics to support Equations (\ref{eqn:predict1}) and (\ref{eqn:predict2}). These statistics are computed for the Stanford Reader.}
\label{table: statistics}
\end{table}

\section{Pointer Annotation Readers}
\label{sec:reference}
In this section, we propose a novel approach, one-hot pointer annotation,  to locate entities in a passage instead of anonymized entity identifiers in the CNN/Daily Mail dataset. In this approach, we use a non-anonymized dataset (WDW), and add a one-hot indicator to each input (word embedding) that indicates occurrences of candidate answers in a passage. This approach simply provides the reference information $R(a,p)$ without losing any information in the passage, unlike anonymized entity identifiers that remove original tokens in the passage.

Additionally, we hope that the one-hot indicator helps aggregation readers that are apparently benefited by the anonymization. The performance of aggregation and explicit reference readers on WDW is in Table (\ref{table:resultswdw}). In the table, the Stanford Reader achieves just better than $45\%$ on WDW while the Attention Sum Reader can get near $60\%$. On the other hand, the performance of the Stanford Reader jumps to near $65\%$ when we anonymize WDW and then re-train the reader. This jump might be explained by the output embeddings $e_o(a)$ to be learned. The output embeddings are semantic word embeddings when the dataset is non-anonymized, but they are semantic-free entity identifiers when the dataset is anonymized. This suppression of semantics may facilitate the separation of the hidden state vector space $H$ into a direct sum $S \oplus E$ with $s(\Phi) \in S$ and  $e_o(a),s(a) \in E$.

\noindent \textbf{One-Hot Pointer Reader.} Here, we implement the one-hot pointer to the Stanford Reader. We modify the input embedding and the output softmax of the Stanford Reader. For the input embedding of a passage, let $i_t$ be the index of a candidate answer in the choice list if the candidate answer is referred to the $t$-th token in the passage, otherwise zero. We define an one-hot pointer $e'(i_t)$ as an one-hot vector of the index if $i_t \neq 0$, otherwise the zero vector, i.e., $e'(0)=0$. Note that a passage in WDW has at most five candidate answers, and we can use a five-dimensional one-hot vector to represent them. Then, we concatenate $e'(i_t)$ as additional features to the word embedding $e(w_t)$ for token $w_t$ in the passage:
\begin{equation}\label{eqn:pointer}
  \bar{e}(w_t) = [e(w_t),e'(i_t)].
\end{equation} 
Then, we replace the input embedding $e(w_t)$ with $\bar{e}(w_t)$ in the Stanford Reader.
For the output softmax, we take the output softmax over some elements of $o$ instead of all elements as follows:
\begin{equation} \label{eqn:answer}
  p(i|d,q) = \softmax_{i \in \mathcal{A}}\; [0, e'(i)]^{\top} o ,
\end{equation}
where ``$0$'' stands for a sufficient number of zeroes in order to make the dimensions match and $o$ is computed by Equation (\ref{eqn:r}). 

Even though not shown here, in preliminary experiments, we also tried a fixed set of ``pointer vectors''\----vectors distributed widely on the unit sphere so that for $i \not = j$ we have that $e'(i)^\top e'(j)$ is small\----instead of one-hot vectors in a case where a choice list has a large number of candidate answers. This reader yields similar performance to the one hot pointer reader while permitting smaller embedding dimensionality.

\noindent \textbf{Linguistic Features.} We also add linguistic features to each input embeddings; whether the current token occurs in the question; the frequency of the current token in the passage; the position of the token's first occurrence in the passage as a percentage of the passage length; and whether the text surrounding the token matches the text surrounding the placeholder in the question. 

Table~\ref{table:resultswdw} shows results when adding these features to the Gated-Attention Reader, Stanford Reader, and One-Hot Pointer Reader, showing large improvements to all readers and leading to the best single-model performance reported on WDW.

\begin{table}
\begin{center}
\begin{tabular}{@{}|l|c|c}
  \toprule
Who-did-What  & \text{Validation (\%)} & \text{Test (\%)} \\
  \hline
Attention Sum Reader        &59.8 &58.8 \\ 
Gated-Attention Reader     &60.3 &59.6  \\
NSE     & 66.5 & 66.2 \\ 
Gated-Attention + Linguistic Features$^{+}$ & 72.2 & \bf{72.8} \\ \hline
Stanford Reader & 46.1 & 45.8 \\
Attentive Reader with Anonymization        &55.7 & 55.5 \\ 
Stanford Reader with Anonymization     & 64.8 & 64.5 \\ 
One-Hot Pointer Reader & 65.1 & 64.4 \\
One-Hot Pointer Reader + Linguistic Features$^{+}$ & 69.3 & 68.7 \\
Stanford with Anonymization + Linguistic Features$^{+}$    & 69.7 &  \textcolor{blue}{\textbf{69.2}} \\ \hline
  Human Performance        &- &84 \\ \bottomrule 
\end{tabular}
\end{center}
\caption{Accuracy on Who-did-What dataset. Each result is based on a single model. Results for neural readers other than NSE are based on replications of those systems.  All models were trained on the relaxed training set which uniformly yields better performance than the restricted training set. The first group of models are explicit reference models and the second group are aggregation models. $+$ indicates anonymization with better reference identifier. \label{table:resultswdw}}
\end{table}

\section{Discussion}
\label{conclusion}
Our experiments indicate that both explicit reference and aggregation readers benefit greatly from this externally provided reference information. Especially, explicit reference readers rely on reference resolution\----a specification of which phrases in the given passage refer to candidate answers.  Aggregation readers also seem to demonstrate a stronger learning ability in that they essentially learn to mimic explicit reference readers by identifying reference annotation and using it appropriately. This is done most clearly in the pointer reader architectures.  Furthermore, we have argued for, and given experimental evidence for, an interpretation of aggregation readers as learning emergent predication structure\----a factoring of neural representations into a direct sum of a statement (predicate) representation and an entity (argument) representation.

At a very high level, our analysis and experiments support a central role for reference resolution in reading comprehension.  Automating reference resolution in neural models, and demonstrating its value on appropriate datasets, would seem to be an important area for future research.

There is great interest in learning representations for natural language understanding. These neural reading comprehension is such that systems still benefit from externally provided linguistic features, including externally annotated reference resolution. It would be interesting to develop fully automated neural readers that perform as well as readers using externally provided annotations.

\section{Conclusion}
In this work, we claimed and empirically showed that the success of aggregation readers and explicit readers could be explained by Equation (\ref{eqn3}), and the contextual and question embeddings could be decomposed into a property and candidate answer symbol. 
For a given passage and question, an aggregation reader computes a score for each token in the passage, which is an inner product between the contextual embedding of the token and the embedding of the question. Then, the aggregation reader predicts the answer by the sum of all contextual embeddings weighted by the score for each token as Equation (\ref{eqn2}). 
On the other hand, an explicit reference reader used explicit reference information that explicitly gives tokens referring to each candidate answer. For each candidate, the explicit reader computes the sum of scores of tokens referring to the candidate answer as Equation (\ref{eqn4}). 

Finally, we proposed one-hop pointer annotation to helps aggregation readers whose performance indicates that these neural networks are benefited from externally provided linguistic features, including externally annotated reference information.

\chapter{Relation and entity centered reading comprehension} \label{sec:wikihop}

In this work, we apply the externally provided reference information that improved the performance of neural readers in Chapter \ref{sec:work2} to another reading comprehension task focusing on not only entities but also their relations, and propose a novel neural model and training algorithm that memory-efficiently trains the model. We propose a Transformer based model with an explicit reference structure that efficiently captures the global contexts. Although the self-attention layer in Transformer consumes a memory that quadratically scales to the length of the input sequence, we propose a training algorithm whose memory requirement is constant to the length of the sequence. We employed Wikihop to show the performance of the model and the training algorithm. The dataset is a reading comprehension dataset focusing on not only entities but also their relations. We presented studies to find an entity from a passage for a given textual query, i.e., cloze-style reading comprehension, in Chapter \ref{sec:work1} and Chapter \ref{sec:work2}. On the other hand, Wikihop is a reading comprehension task whose query consists of a relation and entity and asks another entity that has the relation to the entity. Our model, trained by the algorithm, achieved the state-of-the-art in Wikihop.

\section{Wikihop dataset}

Wikihop consists of a passage, question, candidate answers, and an answer. Here a question is a tuple of a query entity and relation, and then the answer is another entity that has the relation to the query entity. The task is closely related to the relation extraction tasks, and, unlike cloze-style reading comprehension, the task requires not only finding an entity but also understanding relations in the passage. In addition to that, the dataset also provides anonymized passages that help the reference resolution.

\begin{figure}[t]
  \begin{tabular}{cc}
  \begin{minipage}{0.50\hsize}
     \begin{center}
        \includegraphics[width=\textwidth]{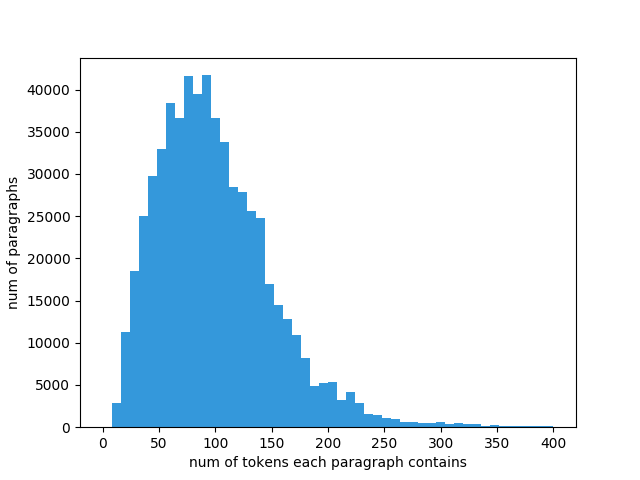}
     \caption{The length of each paragraph in Wikihop.}
     \label{fig:wikihop.paras}
     \end{center}
  \end{minipage} &
  \begin{minipage}{0.50\hsize}
     \begin{center}
        \includegraphics[width=\textwidth]{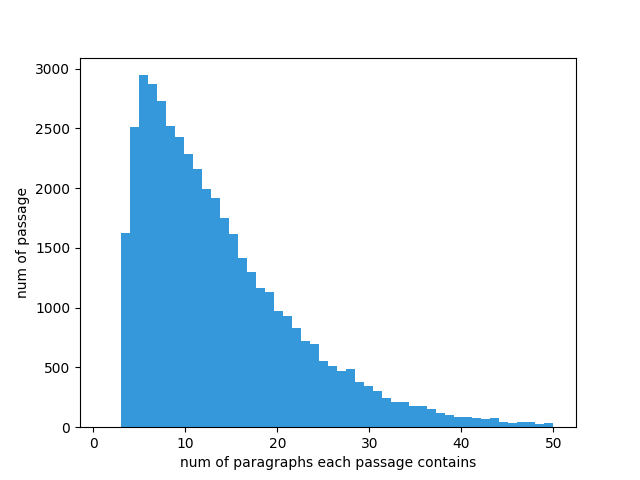}
     \caption{The number of paragraphs for each passage in Wikihop.}
     \label{fig:wikihop.tokens}
     \end{center}
  \end{minipage} 
  \end{tabular}
\end{figure}

Wikihop is designed for multi-hop reading comprehension with relatively long passages. In Wikihop, each passage has multiple paragraphs, as shown in Fig.~\ref{table:wikihop_sample}. In this example the question asks in what country the Hanging Gardens of Mumbai are. Paragraph1 says that the Hanging Gardens of Mumbai are gardens located in Mumbai, and Paragraph2 says that Mumbai is located in India that is a country (Mumbai is a capital city of India). Either of these paragraphs is not enough to infer the answer, India, but both paragraphs are required to infer it. Thus such questions require reading comprehension systems to solve semantic relations over the entire passage, including coreference and inference that is likely difficult to solve. Naturally, the passage consisting of multiple passages is relatively longer than that in other datasets consisting of a single paragraph. Figure \ref{fig:wikihop.paras} and Figure \ref{fig:wikihop.tokens} show the distribution of the number of paragraphs for each passage and the length of each paragraph, respectively. 

\begin{table*}[t]
	\centering
	\fbox{\begin{minipage}[t]{\textwidth}

\textbf{Paragraph1:} The Hanging Gardens, in \underline{Mumbai}, also known as Pherozeshah Mehta Gardens, are terraced gardens … They provide sunset views over the \underline{Arabian Sea} …

\textbf{Paragraph2:} \underline{Mumbai} (also known as Bombay, the official name until 1995) is the capital city of the Indian state of Maharashtra. It is the most populous city in \underline{India} …

\textbf{Paragraph3:} The \underline{Arabian Sea} is a region of the northern Indian Ocean bounded on the north by Pakistan and Iran, on the west by northeastern Somalia and the Arabian Peninsula, and on the east by \underline{India} …

\textbf{Query:} (Hanging gardens of Mumbai, country, ?) 

\textbf{Answer candidates:} \{Iran, India, Pakistan, Somalia, …\}

    \end{minipage}}
    \caption{Sample multi-hop reading comprehension question~\citep{welbl-etal-2018-constructing}.}
    \label{table:wikihop_sample}
\end{table*}

Wikihop is closely related to Wikireading, another relation and entity centered reading comprehension dataset created from Wikipedia and Wikidata. Wikipedia is a free online encyclopedia hosted by the Wikimedia Foundation that consists of more than 6 million articles\footnote{https://en.wikipedia.org/wiki/English\_Wikipedia}. Wikidata is a collaboratively edited knowledge base hosted by the Wikimedia Foundation that is designed as a set of tuples, and each tuple consists of a subject entity, object entity, and their relation. There are more than 7,000 relation types, including ``instance\_of'' and ``location'', and most entities in Wikidata and entries in Wikipedia are linked to each other. Each instance of Wikireading consists of a passage, question, and answer, and it is from a Wikidata tuple, i.e., each question is a relation in the Wikidata tuple, the passage is the Wikipedia article describing the subject entity, and the answer is the object entity.

Wikihop is a reading comprehension dataset constructed from Wikireading, and its passages are carefully selected for multi-hop reading comprehension. The paragraphs are selected on a bipartite graph whose left nodes are entities in Wikidata, and right nodes are paragraphs in Wikipedia. A left entity node is connected to a right paragraph node if and only if its entity is mentioned in the paragraph. Paragraphs on the path between two entities that have a relation in the tuples in Wikidata are used as a passage in a question. The question consists of an entity and the relation on the tuple, and the answer is another entity on the tuple. The paragraphs on the path are used as the passage because the path is likely the reasoning chain to achieve the relation between the two entities. Additionally, unlike Wikireading, Wikihop provides a list of candidate answers for each question that helps to avoid the ambiguity of the answer. 
Thus, Wikihop provides questions that likely require multi-hop reading comprehension, where their answers are inferred from multiple sentences in the passage. 

\section{Related work}
In this section, we review related work for Wikihop by using three approaches. In the first approach, models have various self-attention structures. A limitation of the naive self-attention layer is the maximum length of a sequence that it can take. These models modified the self-attention structure to overcome the limitation; however, their training time (including pre-training and fine-tuning for a downstream task) is longer than those of other models. In the second approach, models consist of a pre-trained encoder and additional network structure, so that they are solely fine-tuned for a downstream task. We also take the pre-training and fine-tuning approach, but we propose a simpler model on the top of an encoder. In the third approach, models are full scratch models whose parameters are all randomly initialized and optimized only on the dataset of the downstream task. These models have no access to the additional linguistic resources used in pre-training and do not perform as well as pre-trained models. 

\subsubsection{Models modifying self-attention structure:} \label{sec:wikihop.transformer_related_models}
In recent years, pre-trained Transformers are surpassing the performance of other neural structures like recurrent neural networks, and convolutional neural networks in reading comprehension tasks. Transformer is a neural structure that processes a sequence by stacked self-attention layers~\citep{VaswaniAttentionNeed}. Each self-attention layer computes an attention from a token to other tokens as follows: 
\begin{equation} \label{eq:selfatt}
    \textrm{Attention}(Q,K,V) = \textrm{softmax}\left( \frac{QK^\top}{\sqrt{d_k}} \right) V ,
\end{equation}
where $Q$, $K$ and $V$ are query, key and value vectors for each token. The network structure is completely geometry free, i.e., there is no structure to reserve the order of tokens in the sequence like recurrent networks, but Transformer takes a position embedding along with a word embedding for each token. This self-attention mechanism gives a rich expressive power to Transformer. 

However, the structure requires an amount of memory that is quadratic in the sequence length in training. The self-attention structure is trained by a stochastic algorithm. The algorithm has two steps to update parameters in the structure. The first step is the forwarding process, where the structure computes the loss through the query, key, and value embeddings. The second step is the backpropagation, where we compute the gradient for each parameter using the query, key, and value embeddings. Thus, the query, key, and value embeddings must be kept until the backpropagation. As Equation (\ref{eq:selfatt}) shows, these embeddings scales quadratically with the sequence length.

Additionally, a pre-trained Transformer has a limitation on the maximum length of sequences that can be processed due to the number of pre-defined position embeddings. The self-attention structure of Transformer does not have any mechanisms that specify the position of tokens except the position embeddings. A position embedding is a trainable parameter, and each position embedding and a corresponding token embedding are paired and injected into the self-attention layer. Again, the self-attention layer has a geometry free structure; thus, the position embeddings are only geometrical information that Transformer can take. In pre-training, a specific number of position embeddings are used; however, the number might not be enough for some downstream tasks where the pre-trained Transformer needs to read longer sequences.

Here, we review approaches that modify the structure, self-attention layer to address the issues. Dynamic self-attention~\citep{zhuang-wang-2019-token} is a self-attention layer whose attention is over top-K tokens selected by a convolutional layer~\cite{8099678}. Transformer-XL and XLNet~\citep{NIPS2019_8812, dai-etal-2019-transformer} have a self-attention layer that uses relative position embeddings rather than absolute positions. A relative position provides the distance between two tokens; a token that we compute the attention from, and another token that we compute the attention to. Thus they are not limited by the number of pre-trained position embeddings. 
Reformer~\citep{Kitaev2020Reformer:} introduced locality sensitive hashing to compute the attention. The locality sensitive hashing provides a subset of all tokens in the sequence that likely dominates the attention score. Thus Reformer reduces the quadratic computational complexity. Longformer~\citep{beltagy2020longformer} employs the idea of a convolutional network where each convolutional unit takes only tokens around it. As the convolutional unit does, Longformer computes attentions for each token over several tokens around it. Additionally, Longformer computes a global attention (attention over all tokens in the sequence) for some special tokens so that they claim the global attention helps to take account of a global context and long dependency. 

Although these approaches potentially solve the fundamental limitation of the Transformer encoder, these models need to be pre-trained from scratch. Typically, these Transformer encoders are pre-trained on a large training data that is much larger than the training data of downstream tasks. As the result, the pre-training is the most time-consuming part of its parameter optimization. Thus, other approaches that are reviewed in the following section add additional structure on the top of pre-trained encoders so that they can avoid the pre-training. 

\subsubsection{Fine-tuning models:}
Another approach is fine-tuning based on pre-trained encoders. In this approach, a model consists of an encoder whose parameters are pre-trained and an additional neural structure whose parameters are randomly initialized. The pre-trained encoder provides contextual word embeddings for each input text. The encoder is pre-trained on a large scale language resource so that it is believed that the encoder obtained some general linguistic knowledge and its contextual word embeddings help downstream tasks. The additional structure is a task-specific neural structure that can efficiently leverage these contextual word embeddings for the downstream task. Thus, the parameters of the structures are fine-tuned for the task during the model is trained on the downstream dataset.
For example, Graph Convolutional Networks is used on the top of Embeddings from Language Model (ELMo) encoder~\citep{de-cao-etal-2019-question, Peters:2018}. \citet{chen2019multihop} proposed a two-stage approach. In the first stage, a pointer network~\citep{NIPS2015_5866} selects a part of the passage that is likely essential for solving the question. In the second stage, a Transformer model takes the part of the passage and finds the answer.

\subsubsection{Other network structures trained from scratch:}
It is worth mentioning that, in some studies, models are trained from scratch. These models consist of a relatively simple encoder and a relatively complicated additional neural structure. For example, \citet{zhong2018coarsegrain} proposed a Coarse-grain Fine-grain Coattention Network consisting of attention over candidate entities mentioned in each paragraph and another attention over the paragraphs on the top of a bidirectional Gated Recurrent Unit (GRU) encoder~\citep{Cho2014LearningTranslation}. \citet{tu-etal-2019-multi} proposed a Heterogeneous Document-Entity (HDE) graph whose node is each entity-mention and paragraph encoded by GRU. \citet{dhingra-etal-2018-neural} proposed a GRU with additional connections between tokens if these tokens are referring to the same entity (coreference).

We propose a simpler and efficient structure that adds a sum layer on the top of a Transformer encoder. Our model works without the time-consuming pre-training, and also our experiments indicate our simple structure efficiently leverages the context embeddings given by the pre-trained Transformer encoder.

\section{Explicit reference transformer}

\begin{figure}[t]
	\centering
	\includegraphics[width=0.80\textwidth]{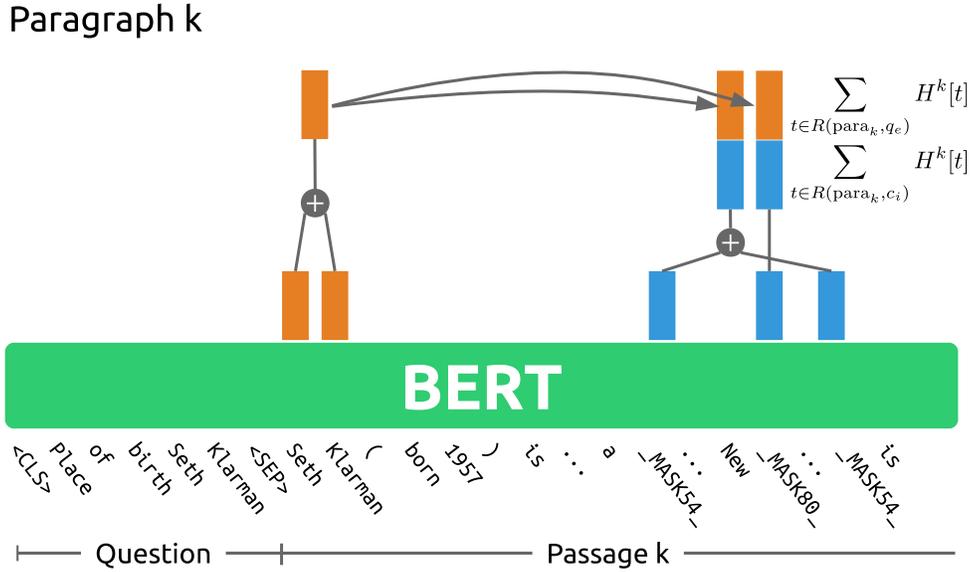}
	\caption{Explicit reference on the Transformer encoder.}
    \label{fig:wikihop.tansformer_explicit}
\end{figure}

We propose a Transformer-based model with the explicit reference structure and a training algorithm for it. Here, the Transformer encoder is a function that takes a sequence of tokens and returns a contextual embedding for each token in the sequence. As we explained in Section \ref{sec:work2}, the explicit reference structure is a neural network structure that explicitly takes the contextual embedding of a token referring to a candidate answer to score the candidate, and these models explicitly leverage these embeddings. 
In this model, the Transformer encoder encodes each paragraph and computes the contextual embeddings of tokens for each paragraph independently, so that its memory usage is linear to the number of the paragraphs and does not quadratically scale with the length of the passage, as we see in Section \ref{sec:wikihop.transformer_related_models}. Then the model accumulates these embeddings over paragraphs and scores the candidate answers. The overview of this model is shown in Figure \ref{fig:wikihop.tansformer_explicit}. 
We also propose a training algorithm for it, which reduces the memory usage during the training to the constant to the number of paragraphs. 

Remembering that the passage is a set of paragraphs, the Transformer encoder encodes the paragraphs independently. We denote the $k$-th paragraph by $\textrm{para}_k$, the question by $q$, and then the encoder parameters by $\Phi$. Then letting the contextual embeddings of the $k$-th paragraph be $H^k$, 
\begin{equation}
	H^k_\Phi = \textrm{Encode}([q, \textrm{para}_k]; \Phi).
\end{equation}
Here the Transformer encoder takes a concatenation of the question and paragraph. The contextual embeddings of a token referring to each candidate answer are accumulated over all paragraphs.

Remembering that each question consists of a relation and entity $q_e$, we also similarly accumulate a query entity embedding, then the candidate answer embeddings are concatenated to the query entity embeddings. Letting the score of the $i$-th candidate answer be $s(c_i)$, then
\begin{equation}
	s(c_i; \Phi) = \theta^\top f \left( \sum_{k} \left[ \sum_{t \in R(\textrm{para}_k, q_e)} H^k_\Phi[t], \sum_{t \in R(\textrm{para}_k, c_i)} H^k_\Phi[t] \right] \right), 
\end{equation}
where $H^k_\Phi[t]$ is the $t$-th contextual representation vector for the given paragraph, $f$ is a fully connected layer, and $R(\mathrm{para}_k,c)$ is the set of positions $t$ where the entity $c$ occurs in the paragraph. To find these positions, we matched entities and noun phrases in the passage whose most words match each entity when entities are not anonymized.

\subsubsection{Training algorithm}

We also propose a stochastic gradient algorithm to train this model, whose memory usage is constant to the number of paragraphs as Algorithm \ref{alg:double_forard}. In this model, the Transformer encoder takes each paragraph instead of the entire passage, so the memory usage of the naive stochastic gradient algorithm is quadratic to the length of paragraphs and linear to the number of the paragraphs, which is still too large to fit a GPU memory when the passage has many paragraphs. During the training, the memory is consumed by a computational graph. The computational graph can be viewed as a representation of an objective function and requires memory for each neural output of parameterized functions in the objective function during the training. For example, parameters of a parameterized function $f(x;\theta)$ is updated by the following during the training, 
\begin{eqnarray}
    f(x; \theta) &=& f_1(f_2(x;\theta)) + g(x) ,\\ \nonumber
    \frac{\partial f}{\partial \theta} &=&  f_1'(f_2(x;\theta)) \frac{\partial f_2}{\partial \theta} ,\\ \nonumber
    \theta &\Leftarrow& \theta - \lambda \frac{\partial f}{\partial \theta}. 
\end{eqnarray}
Here the computational graph keeps the output value of the neural $f_2$ in the forwarding propagation until the backpropagation. Our training algorithm computes the forwarding propagation twice for each backpropagation. In the first forwarding propagation, we compute the loss without keeping all neural outputs, and then in the second forward propagation, we compute the same loss with keeping a subset of the neural outputs whose parameters are updated on the upcoming backpropagation.

In the first forwarding propagation, we compute the contextual embedding for each paragraph independently without keeping neural outputs. We denote the embeddings by $H^k_{\Phi'}$ which is computed as
\begin{equation}
    H^k_{\Phi'} = \textrm{Encode}([q, \textrm{para}_k]; \Phi').
\end{equation}
Here we keep the contextual embedding only and remove the left of neural output values.

In the second forwarding propagation, we recompute the contextual embedding for a single paragraph then compute the total loss with keeping neural outputs for the following backpropagation. We denote the contextual embedding of the target paragraph by $H^k_\Phi$, and
\begin{equation}
	 H^k_{\Phi} = \textrm{Encode}([q, \textrm{para}_k]; \Phi).
\end{equation}
Now we sum the contextual embedding of the target paragraph and that of other paragraphs.
\begin{eqnarray} \label{eq:wikihop.sum_emb}
	e(c_i; \Phi, \Phi') =&& \left[ \sum_{t \in R(\textrm{para}_k, q_e)} H^k_\Phi[t], \sum_{t \in R(\textrm{para}_k, c_i)} H^k_\Phi[t] \right] \\ \nonumber
	&& + \sum_{k'\neq k} \left[ \sum_{t \in R(\textrm{para}_{k'}, q_e)} H^{k'}_{\Phi'}[t], \sum_{t \in R(\textrm{para}_{k'}, c_i)} H^{k'}_{\Phi'}[t] \right].
\end{eqnarray}
Then, the total loss for the passage is
\begin{equation} \label{eq:wikihop.bert.subpara.update.loss}
    \mathcal{L}(q,a; \Phi, \Phi') = \log \frac{\exp \theta^\top f \left( e(a; \Phi, \Phi') \right)}{\sum_i \exp \theta^\top f \left( e(c_i;\Phi, \Phi') \right)},
\end{equation}
where $a$ is the correct answer and only neural outputs under $H^k_{\Phi}$ are stored in the computational graph. And then the gradient is computed with respect to $\Phi$ thus $\Phi$ is updated in the backpropagation, i.e,
\begin{equation}
    \Phi \Leftarrow \textrm{update} \left(\frac{\partial \mathcal{L}(q,a; \Phi, \Phi')}{\partial \Phi} \right) .
\end{equation}
The total loss is computed for each paragraph so that all parameters are updated.

\begin{algorithm}[t]
    \caption{Update steps for each question in the training algorithm that performs the forward propagation twice for the backpropagation.}
    \begin{algorithmic}[1]
        \renewcommand{\algorithmicrequire}{\textbf{Input:}}
        \REQUIRE query $q$, paragraphs $p_0, p_1, ...$, candidate answers $c_0,c_1,...$, and answer $a \in \{c_0,c_1,...\}$
        
        \FOR{$\textrm{para}_k \in \textrm{para}_0,\textrm{para}_1,...$}
            \STATE $H^k_{\Phi'} \Leftarrow \textrm{Encode}([q,\textrm{para}_k]; \Phi') $
        \ENDFOR
        
        \FOR{$\textrm{para}_k \in \textrm{para}_0,\textrm{para}_1,...$}
            \STATE $H^k_{\Phi} \Leftarrow \textrm{Encode}([q,\textrm{para}_k]; \Phi) $
            \STATE $\Phi \Leftarrow \textrm{update}(\frac{\partial \mathcal{L}(q,a; \Phi, \Phi')}{\partial \Phi})$
        \ENDFOR

         \textbf{NOTE:} $\textrm{Encode}(\cdot ;\Phi)$ is a parameterized Transformer that encode a sequence of tokens into a sequence of context aware embeddings, whose parameters are denoted by $\Phi$. $\mathcal{L}$ is described in Equation (\ref{eq:wikihop.bert.subpara.update.loss}).
    \end{algorithmic} 
    \label{alg:double_forard}
\end{algorithm}

\section{Experiments}

Our model is mostly initialized by BERT pre-trained model and fine-tuned on anonymized Wikihop. We use the anonymized version and avoid solving the coreference resolution and identifying mentions of each candidate answer by ourselves so that we use the exact same reference information that other systems used.  The encoder of our model is BERT~\citep{Devlin2018BERT:Understanding}, whose parameters are initialized by BERT-base with twelve self-attention layers and 512 position embeddings. BERT-base is a medium-size Transformer encoder whose scale is similar to Longformer-base. Additionally, we assign a randomly initialized unique word embedding for each anonymized entity in passages. Other parameters are randomly initialized. 
Our model is fine-tuned on Wikihop for five epochs. During the fine-tuning, we permutated candidate answers in each passage to avoid over-fitting. We used 10\% dropout~\citep{JMLR:v15:srivastava14a}, warmup~\citep{DBLP:journals/corr/GoyalDGNWKTJH17} over the first 8\% of the training data, and Adam optimizer~\citep{DBLP:journals/corr/KingmaB14} for the parameter optimization. The learning rate is searched from $2 \times 10^{-6}$ upto $2 \times 10^{-4}$. 

\subsection{Main result}
Table \ref{table:wikihop.main_score} shows the performance of each system on the development data and test data. The first four models are trained from scratch, and the following models are pre-trained on large-scale data and then fine-tuned on the Wikihop training data. The table shows that the performance of our system is significantly higher than those of the other systems on the development data. Our system shows more than 2\% higher accuracy than Longformer-base on the development data. Longformer-base and Longformer-large have 12 and 24 layers for each, and our model uses BERT-base with 12 layers; hence its parameter size is similar to that of Longformer-base. In the test data, Longformer-large achieved the highest accuracy; however, our model achieved the best accuracy in the models with its parameter size scale. Additionally, Longformers are trained on non-anonymized data and they can potentially leverage the information of candidate answer names. On the other hand, our model is trained on anonymized data where candidate answers are replaced by entity IDs; thus, it is impossible to leverage the information of candidate answer names. It is also worth noting that models trained on the anonymized training data perform as good as or better on the non-anonymized test data than the anonymized test data because we can always convert the non-anonymized data into the anonymized data. 

\begin{table}[t]
\centering
\begin{tabular}{@{}lrr@{}}
\toprule
System & Dev accuracy (\%) & Test accuracy (\%)\\ \midrule
GA w/C-GRU~\citep{dhingra-etal-2018-neural}    & 56.0 & 59.3 \\
HDE~\citep{tu-etal-2019-multi}                 & 68.1 & 70.9 \\
CFC~\citep{zhong2018coarsegrain}               & *72.1 & 70.6 \\
DynSAN~\citep{zhuang-wang-2019-token}          & 70.1 & 71.4 \\ \hdashline 
Entity-GCN~\citep{de-cao-etal-2019-question}   & *71.6 & 71.2 \\
BERT-Para~\citep{chen2019multihop}             & 72.2 & 76.5 \\
Longformer-base~\citep{beltagy2020longformer}  & 75.0 & - \\ 
Longformer-large~\citep{beltagy2020longformer} & -    & 81.9 \\ \hdashline
Our model & *77.4 &  - \\ \bottomrule
\end{tabular}
\caption{The performance on the development and test data. The performance on the test data is computed by the leader board system of Wikihop. *Training and development data are anonymized. Note that no anonymized test data is provided. }
\label{table:wikihop.main_score}
\end{table}

\subsection{Ablation studies}
In order to better understand the contribution of the explicit reference structure to the performance, we show two upper bound accuracies; a model that reads each paragraph independently, and an oracle model that solely reads paragraphs mentioning the answer. 

The first model scores each candidate answer for each paragraph independently during the training so that the model does not take account of the contexts beyond each paragraph. Thus the model suggests how much the embedding sum of the explicit reference transformer of Equation (\ref{eq:wikihop.sum_emb}) contributes to capturing the contexts beyond each paragraph. 
On the training, similar to the explicit reference Transformer, the model encodes each paragraph by using a Transformer encoder. 
\begin{equation}
    H^k_{\Phi} = \textrm{Encode}([q,\textrm{para}_k]; \Phi).
\end{equation}
Then, each candidate answer is scored for each paragraph independently unlike the explicit reference reader as follows: 
\begin{equation} 
    \mathcal{L}(q,a; \Phi, \Phi') = \sum_k \log \frac{\exp \theta^\top f \left(\left[ \sum_{t \in R(\textrm{para}_k, q_e)} H^k_\Phi[t], \sum_{t \in R(\textrm{para}_k, a)} H^k_\Phi[t]  \right]\right)}{\sum_i \exp \theta^\top f \left(\left[ \sum_{t \in R(\textrm{para}_k, q_e)} H^k_\Phi[t], \sum_{t \in R(\textrm{para}_k, c_i)} H^k_\Phi[t] \right]\right)}.
\end{equation}
The model predicts the answer by the maximum score over the paragraphs, i.e.,
\begin{equation}
    \hat{a} =  \underset{i}{\textrm{argmax}} \max_k \ \theta^\top f \left(\left[ \sum_{t \in R(\textrm{para}_k, q_e)} H^k_\Phi[t], \sum_{t \in R(\textrm{para}_k, a)} H^k_\Phi[t]  \right]\right).
\end{equation}

The first row of Table \ref{table:wikihop.ablation} shows the accuracy of the model. The accuracy dropped by 8\% from our full explicit reference Transformer. This gap indicates that the simple embedding sum significantly contributes to capturing the contexts beyond each paragraph. 

The second model is an oracle model that takes solely paragraphs containing the correct answer so that it gives an identical maximum performance of the explicit reference Transformer in each paragraph. The model is trained and tested solely on paragraphs containing the correct answer. It is worth noting that the oracle is strong and removes most of the candidate answers. 

The second row of Table \ref{table:wikihop.ablation} shows the accuracy of the oracle model. Naturally, the performance is better than those of non-oracle models, and the strong accuracy indicates the potential of the explicit reference Transformer.

\begin{table}[t]
\centering
\begin{tabular}{@{}lr@{}}
\toprule
System & Dev accuracy (\%)\\ \midrule
Independent paragraphs & 69.4 \\
Oracle paragraphs & 96.9 \\ \hdashline
Our model & 77.4  \\ \bottomrule
\end{tabular}
\caption{The model of independent paragraph reads each paragraph independently, and the model of oracle paragraphs takes solely paragraphs mentioning the correct answer.}
\label{table:wikihop.ablation}
\end{table}


\section{Conclusion}
We proposed the explicit reference Transformer that has a simple sum layer on the top of a pre-trained Transformer encoder. The sum layer, called explicit reference structure, performs over contextual token embeddings referring to each candidate answer. Our model is simple and efficiently fine-tuned over Wikihop, and its performance is significantly better than that of models with the similar parameter size.

We also proposed a novel stochastic gradient descent training algorithm. The algorithm performs the forward computation twice; one for computing contextual embeddings and another for storing all neural outputs for the backpropagation. The algorithm requires a constant size of the memory-usage to the length of the input text; thus, it memory-efficiently trains the Transformer encoder. 

For future work, we would like to apply this model to other datasets to show the robustness of this approach. The Transformer encoder encodes geometric information by solely position embeddings, unlike recurrent networks and convolutional networks that encode geometric information by their network structures. However, the Transformer encoder, we believe, strongly associated with the geometry of the input sequence, and the contextual token embedding on the top of the $t$-th token is mostly representing the token. Hence, using the explicit embeddings of a task-specific token seems a promising approach. 

\chapter{Relation extraction with weakly supervised learning for materials science}  \label{sec:work3}

In this chapter, we present our work in relation extraction for materials science~\citep{doi:10.1080/14686996.2018.1500852}. As we described in Section \ref{sec:entity_relation}, relation extraction is studied in the context of knowledge base population, however; it can be view as a reading comprehension desiring a relation between two given entities. Thus, in this study, we find a relation between two given entities from a text resource, and also we build a graph using the relations that visualize the knowledge described in the text resource. Additionally, this work is collaborative work with materials science, and our target knowledge to be visualized is information that helps material development. 

A key strategy to build the structured knowledge in materials science is Processing-Structure-Property-Performance (PSPP) reciprocity~\citep{weixiong}. The PSPP reciprocity is a framework to understand material development, a field of study to find a manufacturing process that gives a material with specific properties. The PSPP reciprocity explains how the manufacturing process gives a property of a material on three steps: process, structure, and property. The first step is a set of processings where each processing is a (typically) chemical or physical input to the material. The second step is a set of structures where each structure of the material is a pattern of molecules in the material. The third step is a set of properties. Each property is a character of the material that we find valuable. The PSPP reciprocity explains that the first step -- processings -- builds structures in the material, and the second step -- structure -- gives some properties of the material, then the third step -- property -- gives the performance of the material.

\begin{figure}[th]
    \centering
    \includegraphics[width=1.0\textwidth]{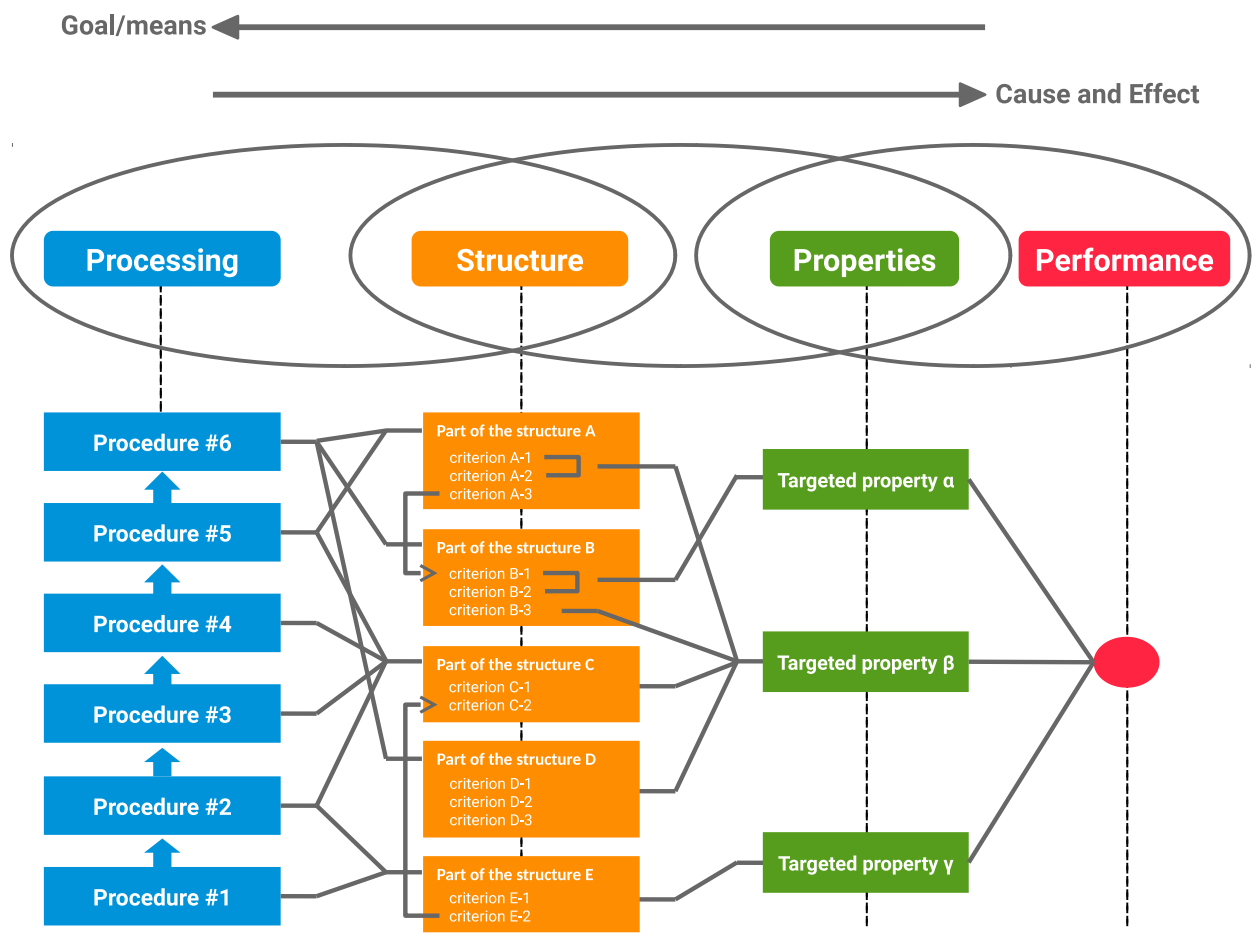}
    \caption{The process-structure-property-performance reciprocity}
    \label{fig:cohen}
\end{figure}

The PSPP reciprocity derives a knowledge graph, and PSPP chart defined as follows. In the knowledge graph, each node represents a specific process, structure, or property, a node of processing has an edge to a node of a structure if the processing builds the structure, and the node of the structure has an edge to the node of a property if the structure affects the property. Then no node of processing and no node of a structure are connected because, according to the PSPP reciprocity, all properties are given by processings through structures. A subset of the knowledge graph is called a PSPP chart, e.g., Fig.\ref{fig:cohen}~\citep{olson2000}. These edges in the PSPP chart indirectly visualize processings that impact on specific desired properties and help material development. 

Even though PSPP charts are practically helpful in material development, there are a huge number of nodes in the knowledge graph, and it is expensive to find all edges by hand. Hundreds of processings, structures, and properties are known in materials science. Thus the number of all possible edge is exponentially large, and finding such a number of edges by hand is practically impossible. In practice, expert researchers draw a PSPP chart, subgraph of the knowledge graph around target properties.

In this research, we developed a computer-aided material design system that automatically finds a PSPP chart from given scientific articles. The system is based on weak supervision that is well studied in the context of knowledge base completion, such as TAC\footnote{https://tac.nist.gov}. Here, the system is trained on about 100 relationships and thousands of non-annotated scientific articles from Elsevier's API\footnote{https://dev.elsevier.com}, and then completes all relations among processing/structure/property nodes. The system does not rely on any specific dataset such as AtomWork~\citep{atomwork}, but it relies on scientific articles that likely cover the knowledge needed to fill the PSPP chart. Then, the system visualizes processings that likely impact on given target properties.

\section{Related work}

This study is closely related to knowledge base population, a task to find relations between entities in a knowledge base. A knowledge base is a well-structured database consisting of relationships among entities, i.e., tuples of an entity-pair and relation. For the knowledge base, it is difficult to complete all relationships in the knowledge base by hand, and automatic approaches to complete the knowledge graph from texts are studied in the field of NLP.

In these approaches, we used distant supervision~\cite{Mintz:Distantsupervisionforrelationextractionwithoutlabeleddata}. In distant supervision, we preprocess the training data; a subgraph of the knowledge base (tuples of a pair of entities and their relation) and corpus (raw text), and then generate weakly labeled sentences. Each weakly labeled sentence is a sentence mentioning multiple entities whose relation is in the subgraph, and labeled by the relation. In other words, the weakly labeled sentence is distantly labeled by the relation on the knowledge base. Then these weakly labeled sentences are used to train machine learning models. The labels of these sentences seem noisier than manual labeling for each sentence, and the noise reduction of these labels is a key to this approach.

Feature-based machine learning models and convolutional neural network (CNN) models are studied in the distant supervised approach. In recent years, CNN models have surpassed feature-based models~\citep{Hoffmann:Knowledge-BasedWeakSupervisionforInformationExtractionofOverlappingRelations, Surdeanu:Multi-instanceMulti-labelLearningforRelationExtraction, Fan:DistantSupervisionforRelationExtractionwithMatrixCompletion, Riedel:ModelingRelationsandTheirMentionswithoutLabeledText}. 
Residual learning is used to help the deep CNN network~\cite{Huang2017DeepResidualLearningforWeakly-SupervisedRelationExtraction}. 
Zeng et al.~\cite{Zeng:DistantSupervisionforRelationExtractionviaPiecewiseConvolutionalNeuralNetworks} split a sentence into three parts, then applied max pooling to each part of the sentence over a CNN layer. 
Sentence level attention is introduced for selecting a key sentence. In this approach, a network takes a set of sentences for a relation between two entities. Each sentence contains both entities. An attention mechanism over a CNN allows the network to automatically select a key sentence which is likely describing the desired relation. It seems helpful to overcome the noise of distant labels~\cite{Lin:NeuralRelationExtractionwithSelectiveAttentionoverInstances, Ji:DistantSupervisionforRelationExtractionwithSentence-LevelAttentionandEntityDescriptions, Liu:ASoft-labelMethodforNoise-tolerantDistantlySupervisedRelationExtraction}.

\section{Preliminary} \label{sec:task}
Our task is to complete a PSPP knowledge graph from scientific articles and extract a subgraph of the PSPP knowledge graph. Let $E$ be the entities of the knowledge graph, and $r_{e_i, e_j} \in \textit{bool}$ be the relation between entities $e_i$ and $e_j$. The subgraph of PSPP knowledge graph is a set of PSPP charts, e.g., $\{(e_i, e_j, r_{e_i, e_j}) | e_i, e_j \in E' \subset E\}$. Here $r_{e_i, e_j} = \textit{True}$ if entities are connected in the chart and $r_{e_i, e_j} = \textit{False}$ otherwise (see Fig. \ref{fig:cohen}). Let $S = \{s_0, s_1, ...\}$ be the sentences in the scientific articles, and then sentences mentioning entities $e_i$ and $e_j$ be $S_{e_i, e_j} \subset S$. In the task, we find all relations among the entities, i.e., $\{r_{e_i,e_j}| \forall \ e_i, e_j \in E\}$.

\section{System description} \label{sec:system_description}
Our system completes the PSPP knowledge graph by two steps; entity collection and relation identification, and then produces a PSPP chart for given properties from the knowledge graph. 

In the first step, our system collects entities $E$ in the PSPP knowledge graph, and then these entities were classified into three material development steps; processing, structure, and property. For example, `tempering' and `hot working' are classified into processing, `grain refining' and `austenite dispersion' are classified into structure, and then `strength' and `cost' are classified into property. 

In the second step, our system identifies relations among entities  $r_{e_i, e_j}$ from scientific articles. Here a machine learning model is trained on weakly labeled sentences, i.e.,
\begin{equation}
 \{(S_{e_i, e_j}, r_{e_i, e_j}) | e_i, e_j \in E_{train} \subset E\} ,
\end{equation}
where $E_{train}$ is a set of entities in PSPP charts for training. The trained model fills other relations to complete the PSPP knowledge graph. 

Then additionally, our system finds and visualizes processes that likely impact on given properties. Here, we assume a scenario where a researcher is developing a new material with certain desired properties and looking for processes related to the properties in a PSPP chart. In this scenario, the PSPP chart is with certain processes and structures around the desired properties.

\subsubsection{Entity collection} \label{sec:factor}
In this section, we describe how we collected entities in the knowledge graph. The entities are collected from two resources; Scripta Materialia\footnote{https://www.journals.elsevier.com/scripta-materialia} and scientific articles.

Scripta Materialia is a journal with a keyword list for helping identify the topic of each article. The keyword list has five sections; 1) Synthesis and Processing; 2) Characterization; 3) Material Type; 4) Properties and Phenomena; and 5) Theory, Computer Simulations, and Modeling. We used keywords in 1) Synthesis and Processing for processing, keywords in 3) Material Type for structure, and keywords in 4) Properties and Phenomena for property.

\begin{figure}[t]
    \centering \includegraphics[width=0.7\textwidth]{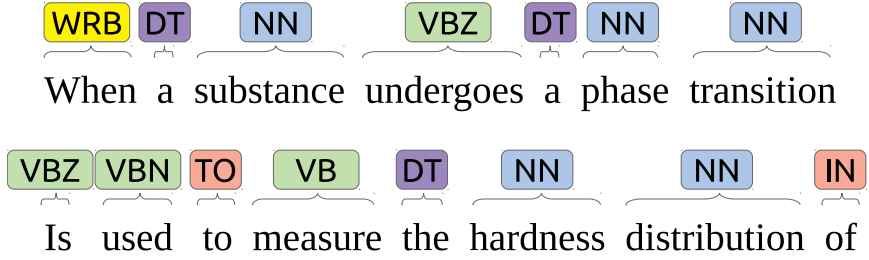}
\caption{Sentences containing noun phrases. }
\label{fig:sentence_with_noun}
\end{figure}

Additional structures are collected from nouns phrases in scientific articles. These noun phrases consisting of multiple NNs (singular nouns, or mass nouns), are collected from a corpus described in Section \ref{sec:corpus} by using Stanford CoreNLP~\citep{manning-EtAl:2014:P14-5}, then  each noun phrase is classified into structure if it does not contain any words in the keyword list. The phrase containing a keyword is classified as the class of the keyword. For instance, Fig. \ref{fig:sentence_with_noun} lists two sentences with noun phrases. Here `phrase\_transition' is classified as a structural entity, but `hardness\_distribution' is classified as a property entity, as `hardness' is in the keyword list. We collected such additional structures because the number of structural entities is significantly greater than those of processing and property entities, and the keyword list is not long enough to cover structural entities from a materials science standpoint.

All keywords and the $n$ most frequent noun phrases are collected, and each word/phrase is assigned a node in the PSPP knowledge graph. The total numbers of entities were 500, 500, and 1000 for process, property, and structural entities respectively. Table \ref{fig:factors} lists samples of the $n$ most frequent phrases.

\begin{table}[h]
\centering
\caption{Samples of entities obtained by the linguistic rules}
\label{fig:factors}
\begin{tabular}{@{}lll@{}}
\toprule
\multicolumn{1}{c}{Process} & \multicolumn{1}{c}{Structure} & \multicolumn{1}{c}{Property} \\ \midrule
water quenching             & carbon dioxide                & creep behavior               \\
element modeling            & grain distribution            & fatigue behavior             \\
peak temperature            & particle size distribution    & misorientation angle         \\
rolling texture             & matrix phase                  & shock resistance             \\
deformation mode            & $\beta$ titanium alloy              & fracture strain              \\
microwave sintering         & $\beta$ grain size                  & tensile ductility            \\
plasma sintering & solution strength & fracture behavior \\
discharge machining         & pore size                     & vacuum induction melting     \\ \bottomrule
\end{tabular}
\end{table}

\subsubsection{Relation identification} \label{sec:model}

\begin{figure}[t]
    \centering
    \includegraphics[width=0.95\linewidth]{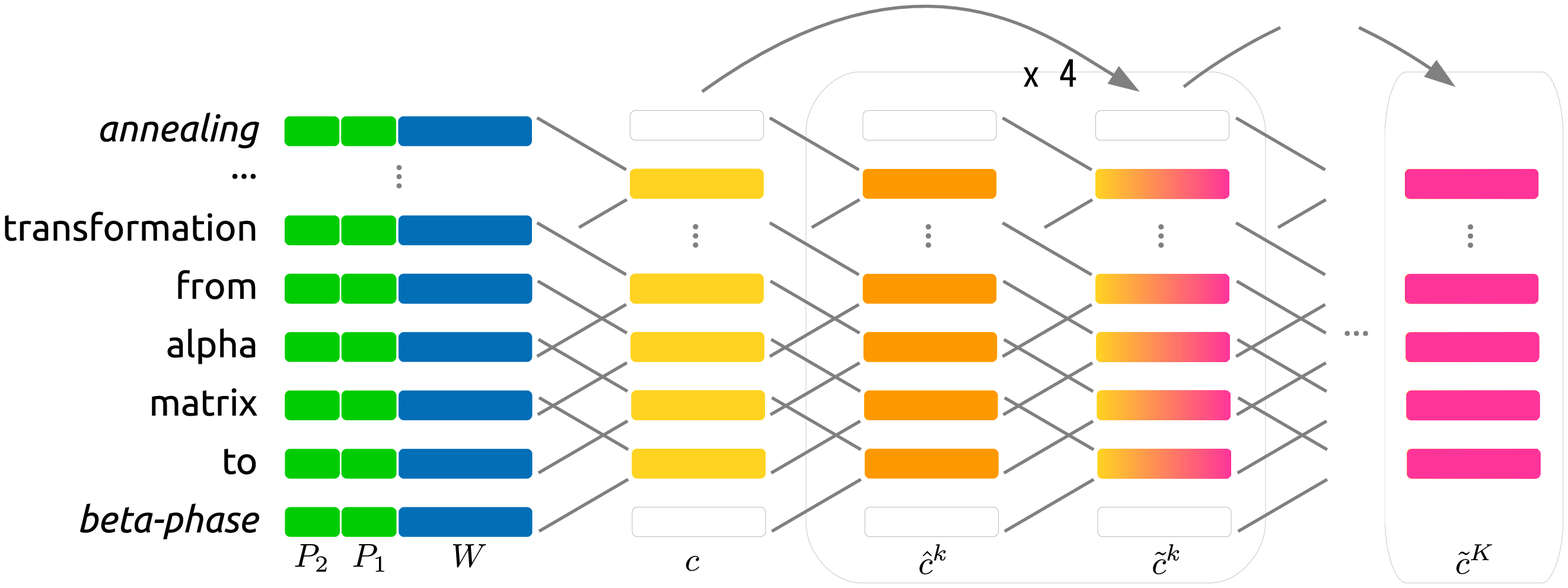}
    
    \vspace{1em}
    \centering
    \includegraphics[width=0.4\linewidth]{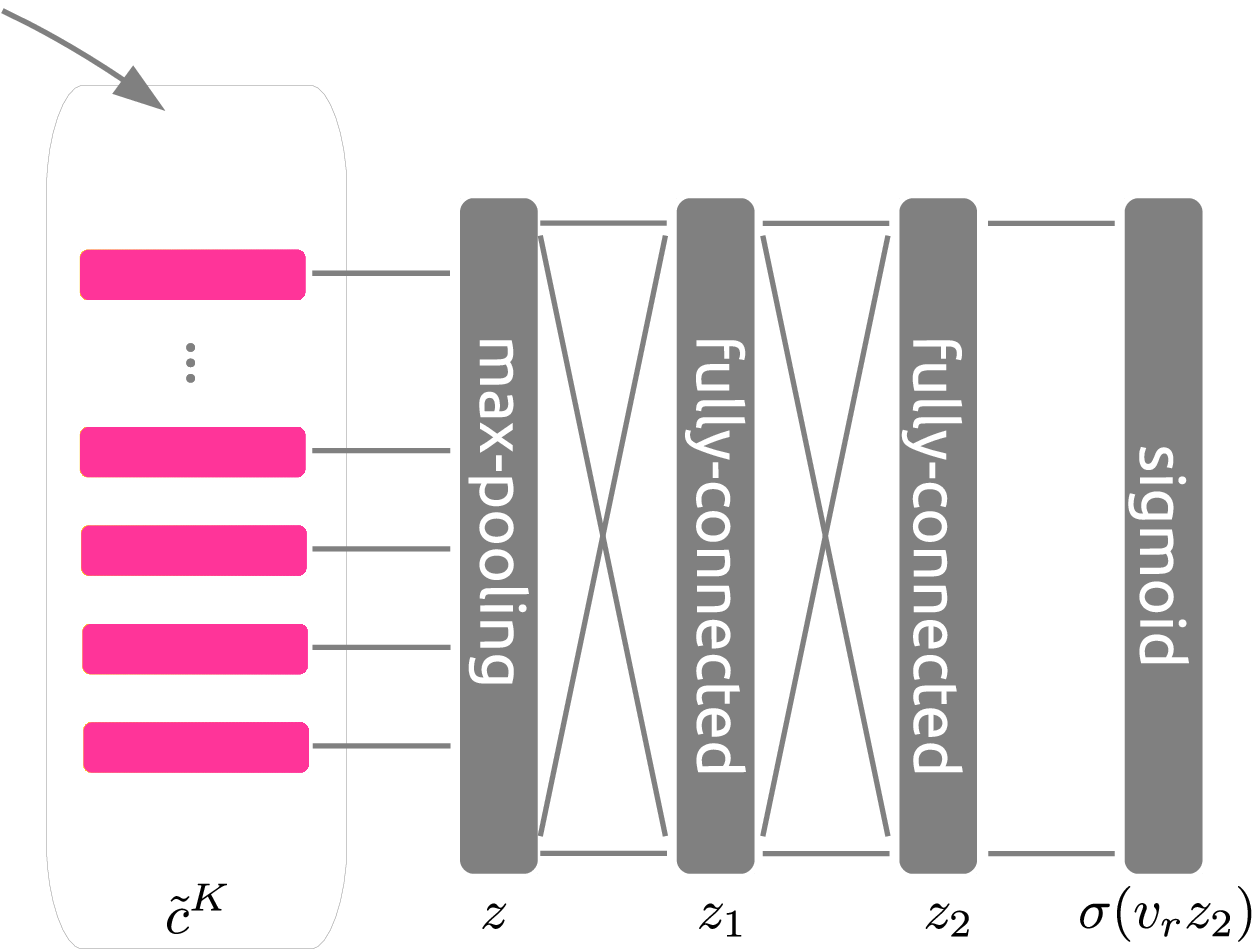}
    \caption{Structure of the CNN model. The convolutional layers embed a sentence, and the max pooling and two fully connected layers give a binary probability distribution with a sigmoid function.}\label{fig:mlp}
\end{figure}

In this section, we describe our CNN model for identifying the relation between entities. We use a stacked CNN with residual connections~\citep{Huang2017DeepResidualLearningforWeakly-SupervisedRelationExtraction}. The CNN model consists of convolutional units with a deep residual learning framework that embeds the sentence into a vector representation. Then, the vector representation produces the probability distribution of the binary relation with a sigmoid layer. We show the overview of the model in Fig. \ref{fig:mlp}.

The CNN model takes each weakly labeled sentence. Let the sentence be  $s = \{t_0, ..., t_i,...\}$, where $t_i$ is the $i$-th token in the sentence, and $W(t_i) \in \mathcal{R}^{d_w}$ be a word embedding of the token $t_i$. We define a relative distance from a token to an entity in the sentence as $k-i$ where $k$ is the position of the entity and $i$ is the position of the token. Let the relative position embedding of the token be $P(k-i) \in  \mathcal{R}^{d_p}$. We define the token embedding as
\begin{equation}
    x_i = [ W(t_i); P_1(k_1 - i); P_2(k_2 -i )],
\end{equation}
where $k_1$ and $k_2$ are the first and second entity in the sentence. 
Note that each sentence $s$ is padded to a fixed length $L$, and any relative distance greater than $D_{\textrm{max}}$, is treated as $D_{\textrm{max}}$. 

We put the token embeddings into the first convolutional layer. The convolutional unit of the first layer takes token embeddings around the position $i$, and computes  $c_i \in \mathcal{R}^{d_c}$ as follows:
\begin{equation}
    c_i = g(\textbf{w} x_{i:i+h} + b) ,
\end{equation}
where $x_{i:i+h}=[x_i;x_{i+1};...;x_{i+h-1}]$, $\textbf{w} \in \mathcal{R}^{d_c \times h(d_w + 2d_p)}$ and $b \in \mathcal{R}^{d_c}$ is a bias. $g$ is an element-wise non-linear function, ReLU.

Following the first convolutional layer, the other convolutional layers are stacked with residual learning connections that directly transmit a signal from a lower to a higher layer while skipping the middle layers. We define these two adjacent convolutional layers called a residual CNN block as follows:
\begin{eqnarray}
    \hat{c}^k_i &=& g(\hat{\textbf{w}}_k (\tilde{c}^{k-1}_{i:i+h} + \tilde{c}^{k-2}_{i:i+h}) + \hat{b}_k) ,\\
    \tilde{c}^k_i &=& g(\tilde{\textbf{w}}_k \hat{c}^k_{i:i+h} + \tilde{b}_k) ,
\end{eqnarray}
where $\tilde{c}^0 = c$. Here, the first convolutional layer $\hat{c}^k_i$ takes a signal from the immediately lower layer $\tilde{c}^{k-1}_{i:i+h}$ and another signal from the lower block $\tilde{c}^{k-2}_{i:i+h}$. 

We put the output of the last convolutional layer into a max pooling layer. Denoting the last output as $\tilde{c}^K \in \mathcal{R}^{ L-h+1 \times d_c}$, 
\begin{equation}
    z = \underset{i}{\textrm{maxpool}} \ \tilde{c}^K_{i} .
\end{equation}
Then, we put $z$ into two fully connected layers and a sigmoid function that gives the probability distribution of the desired relation given the sentence $P( r  | s )$:
\begin{eqnarray}
    z_1 &=& g( \textbf{w}^{g_1} z + b^{g_1} ) , \\
    z_2 &=& g( \textbf{w}^{g_2} z_1 + b^{g_2} ) ,
\end{eqnarray}
\begin{equation}
     P( r = \textrm{True} | s ) = \sigma( v_r z_2 ) ,
\end{equation}
where $r$ is the binary relation between the entities, $\textbf{w}^g \in \mathcal{R}^{d_c \times d_c}$ and $b^g \in \mathcal{R}^{d_c}$. 

The desired probability $P( r = \textrm{True} | e_i, e_j )$ is the maximum of the probabilities over sentences. This is
\begin{equation}
    P( r = \textrm{True} | e_i, e_j )  = \max_{s \in S_{e_i, e_j}} P( r = \textrm{True} | s ).
\end{equation}

The model is trained on a naive distant supervised approach, where the objective function is maximized for each sentence,
\begin{equation}
    \underset{\Phi}{\textrm{max}} \sum_{ (e_i, e_j) \in E_{\textrm{train}}} \sum_{s \in S_{e_i, e_j}} \log P( r_{e_i, e_j} | s ) ,
\end{equation}
where the parameters $\Phi = \{W , P_1, P_2, \textbf{w}, b\}$.

\subsubsection{Branching} \label{sec:graph_completion}
Additionally, we generate a PSPP chart from the knowledge graph for given desired properties. Here the PSPP chart is a subgraph of the knowledge graph that indicates processings that are likely impact on the desired properties. We find the PSPP chart by considering a max-flow problem where the flow occurs from the given properties to the processings. The inlets are all processings and the outlets are given properties. The capacity of each edge is the score of the relation, i.e., $P(r=\textrm{True}|e_i, e_j)$. We maximize the amount of flow with a limited number of nodes in the graph.

We compute the capacity of each entity in the graph, which is the amount of flow that it can accept. Recalling that nodes of structure are connected to property and processing, and no processing and property are connected, all flows pass through nodes of structure. We define the capacity of  a node of structure $e$ as
\begin{equation}
    C(e) = \textrm{min} \left( \sum_{e' \in \textrm{PRC}} P( r = \textrm{True} | e, e' ), \sum_{e \in \textrm{PRP}'} P( r = \textrm{True} | e',e ) \right),
\end{equation}
where PRC is a set of all nodes of processing and PRP' is a set of the desired properties. Similarly, we define the capacity of processing as
\begin{equation}
    C(e) = \sum_{e' \in \textrm{STR}'} P( r = \textrm{True} | e, e' ) 
\end{equation}
where STR' is a set of all nodes of structure.

The produced PSPP chart is composed of $n$ processings, $m$ structures, and the desired properties where $n$ and $m$ are the given hyper-parameters. The entities of the processing/structure are the $n$ and $m$ most capable nodes. For efficiency, the nodes are greedily searched so that optimality is not guaranteed. The PSPP chart shows the processings/structures related to the desired properties.

\section{Experiment for relation identification} \label{sec:experiment}

The CNN model in Section \ref{sec:model} was trained and evaluated on a set of PSPP charts and scientific articles. The model was trained on weakly labeled sentences mentioning entities in PSPP charts for training, and then it took weakly labeled sentences mentioning entities in held-out PSPP charts for testing and predicted relations between entities in these held-out PSPP charts. In both of training and testing, the weakly labeled sentences are found in the scientific articles in Section \ref{sec:corpus}.

\subsubsection{PSPP charts}\label{sec:relation_data}
We used four PSPP charts~\citep{weixiong} for training and testing. These four charts have 104 entity pairs in total as shown in Table \ref{table:factors_data} and Table \ref{table:relations_data}. We used three arbitrary charts for training and the fourth chart for testing. Thus, we trained and tested our model on four pairs of training and test charts. We used the likelihoods of relationships in these four test charts for computing precision and recall curves in Section \ref{sec:material:eval} to obtain a smooth curve.

\begin{table}[t]
\centering
\caption{Entities in the relationship data}
\label{table:factors_data}
\begin{tabular}{l|r}
\hline
Category & Size \\ \hline
Process  &    17     \\
Structure &   21     \\
Property  &   6      \\
\hline
\end{tabular}
\end{table}

\begin{table}[t]
\centering
\caption{Relations in the relationship data}
\label{table:relations_data}
\begin{tabular}{l|rr}
\hline
Relationship type & Positive & Negative \\ \hline
Process $\leftrightarrow$ Structure &  14 & 49 \\
Structure $\leftrightarrow$ Property & 10 & 31 \\
\hline
\end{tabular}
\end{table}

\subsubsection{Scientific articles} \label{sec:corpus}
We used publicly accessible scientific articles on ScienceDirect\footnote{https://www.sciencedirect.com} for training and testing. ScienceDirect is an Elsevier platform providing access to articles in journals in a variety of fields, such as social sciences and engineering. Approximately 3,400 articles were collected using the keyword (`material' and  `microstructure') on ScienceDirect, i.e., each article is related to both `material' and `microstructure'. About 5,000 weakly labeled sentences were founded in these scientific articles by using the four PSPP charts, i.e., roughly 50 sentences for each entity pair on average.

\subsubsection{Training detail}
We trained our CNN model described in Section \ref{sec:model} on weakly labeled sentences. Each weakly labeled sentence is labeled as follows. Let a set of sentences mentioning entities $e_i$ and $e_j$ be $S_{e_i, e_j}$. Here each entity is mapped to a span in a sentence by max-span string matching, i.e., an entity is mapped to the span if the span is the entity name, and no other entity names overlap it. For instance, 
\begin{itemize}
    \item Within each \textbf{phase}, the properties are ...
    \item When a substance undergoes a \textbf{phase} transition ...
\end{itemize}
The \textbf{phase} in the first sentence is mapped to entity `phase', but \textbf{phrase transition} is mapped to `phase\_transition' instead of `phase' in the second sentence. Thus a sentence mentions an entity if and only if it is mapped on a span in the sentence. 

The model parameters are optimized by stochastic gradient descent and dropout. Dropout randomly drops some signals in the network that are thought to help the generalization capabilities of the network. We employed an Adam optimizer with a learning rate of 0.00005, and randomly dropped signals from max pooling during training with a probability of 20\%. The word embeddings were initialized with
GloVe vectors~\cite{Pennington2014GloVe:Representation}. Other hyper-parameters are listed in Table \ref{table:params}. 

\begin{table}[t]
\centering
\caption{Hyper-parameters of the CNN model}
\label{table:params}
\begin{tabular}{c|r}
\hline
Parameter &  Value  \\ \hline
$L$ & 100 \\
$D_{\textrm{max}}$ & 30 \\
$K$ 			& 4 \\
$h$				& 2 \\ 
$d_w$           & 50  \\
$d_p$           & 5  \\
$d_c$ 			& 50 \\
$L_2$ regularization & 0.0001 \\
\hline
\end{tabular}
\end{table}

\subsubsection{Baseline models} \label{sec:baseline}
\begin{figure}[t]
	\centering
	\includegraphics[width=0.80\textwidth]{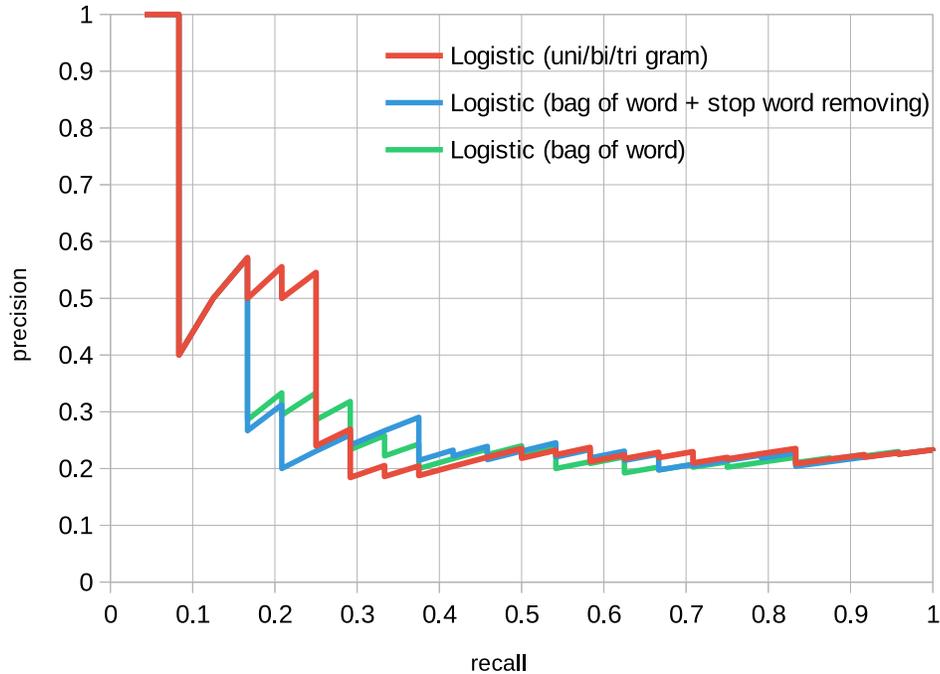}
	\caption{Precision-recall curve of the logistic regression model. The features are `bag of words', `bag of words + stop word removal' and `bag of unigram + bigram + trigram'}
    \label{fig:logistic}
\end{figure}

\begin{figure}[t]
	\centering
	\includegraphics[width=0.80\textwidth]{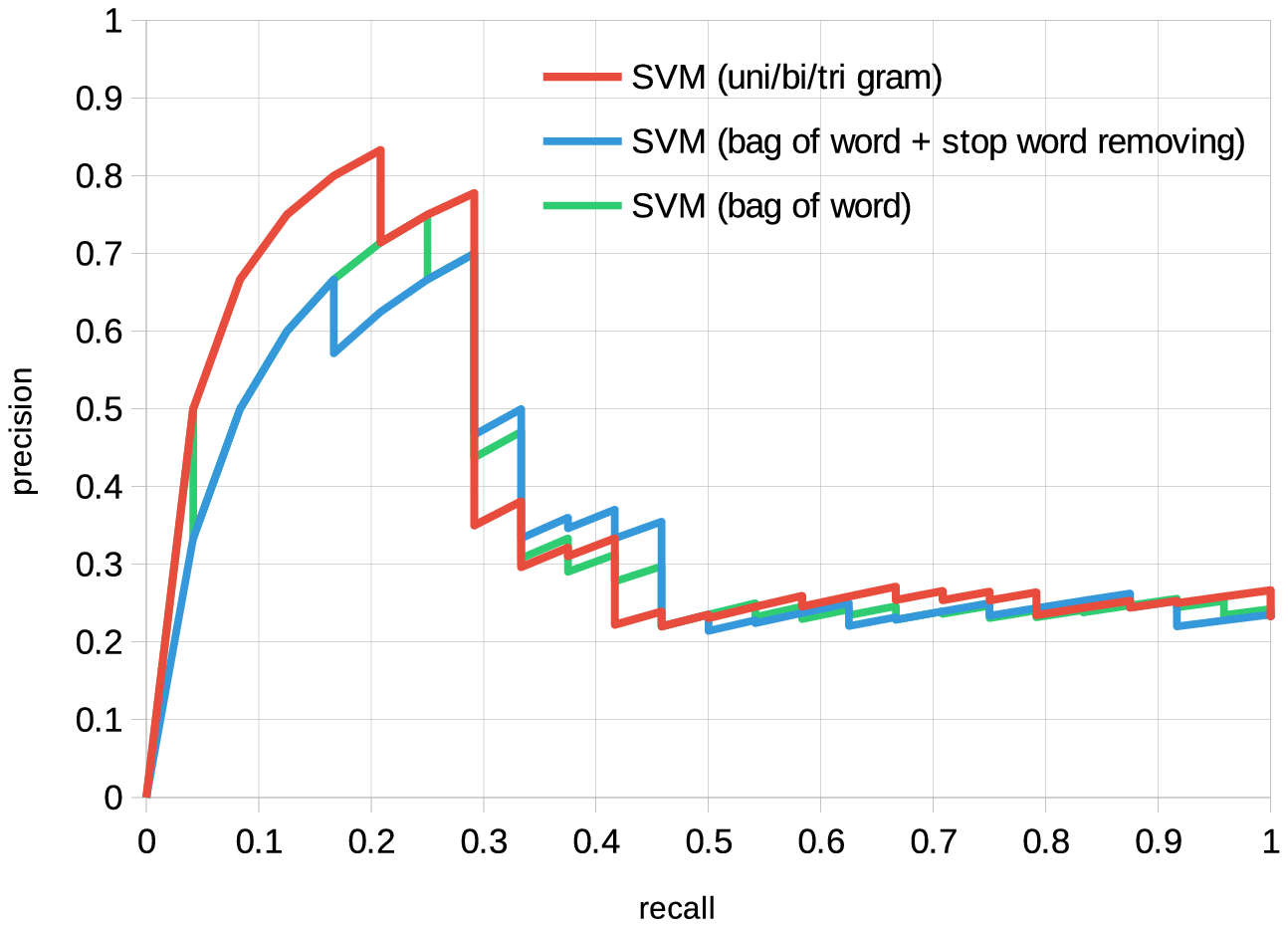}
	\caption{Precision-recall curve of the SVM model. The features are `bag of words', `bag of words + stop word removal' and `bag of unigram + bigram + trigram'}
    \label{fig:svm}
\end{figure}

We compared the performance of our CNN model to the performance of legacy machine learning models; Logistic regression and Support Vector Machine (SVM). The baseline models are trained on weakly labeled sentences and predicted a binary relation for a given entity pair as the CNN model did. The models used the bag-of-words feature that indicates whether a word is in a set of sentences. The feature is represented by a sparse binary vector, where an element is one if the corresponding word is in the sentences and zero otherwise. We also explored stop words removal and n-gram features in Fig. \ref{fig:logistic} and Fig. \ref{fig:svm}; however, the effect was limited. Note that the radial basis function (RBF) kernel was used in all SVM models. 

\subsubsection{Evaluation metric} \label{sec:material:eval}
The evaluation metrics are precision and recall, which are standard metrics for information extraction tasks. Precision is the ratio of correctly predicted positive entity pairs to all predicted positive entity pairs, and gives the accuracy of the prediction. Recall is the ratio of correct predictions to all positive entity pairs in the test data, and gives the coverage of the prediction. A positive entity pair is a pair whose relation is True. 
We obtain high precision and low recall if a system returns only a small number of high confidence predictions, and low precision and high recall if a system returns many low confidence predictions. Typically, these are balanced by a hyper-parameter (confidence) of system prediction. Thus, the trajectory of precision and recall pairs is computed with various values of the hyper-parameter, and is called a precision-recall curve.

In this evaluation, the hyper-parameter was an integer $t$, the number of positive entity pairs in the prediction. For a given $t$ and a set of entity pairs in the test relationship data, the system predicts a binary relation, for each pair. It predicts the $t$ most likely positive pairs, and the other pairs are predicted as negative. 

The entity pairs in the test relationship data were scored by a machine learning model trained on the corresponding training relationship data, where the score was $P( r = \textrm{True} | e_i, e_j )$. A test data corresponded with a training data, unaware of the relationships in the test data (Section \ref{sec:relation_data}). A model was trained on the corresponding training data and scored a pair in the test data to avoid letting the model know the true relationships during training.

Then, a precision and recall pair for a given hyper-parameter $t$ was computed as follows:
\begin{eqnarray}
    \textrm{Precision}_t &=& \frac{| R_t \cap R_{test} |}{ t } ,\\
    \textrm{Recall}_t &=& \frac{| R_t \cap R_{test} |}{ | R_{test} | } ,
\end{eqnarray}
where $R_{test}$ is the set of entity pairs with positive relations in all test relationship data, and $R_t$ represents the $t$ most likely positive entity pairs. The likelihood was a score given by the model.

\section{Results} \label{sec:result}

Figures \ref{fig:logistic} and \ref{fig:svm} show the precision-recall curves for the baseline models. These figures show various feature representation schemes, such as stop words and n-grams (Section \ref{sec:baseline}) on the logistic and SVM models. The logistic model performed well on low recall space, i.e., most confidently predicted positive entity pairs were actually positively related. On the contrary, the performance of the SVM model was poorer in the space but better overall than the logistic model. In both models, the effects of the feature representation schemes were limited. 

\begin{figure}[t]
	\centering
	\includegraphics[width=0.80\textwidth]{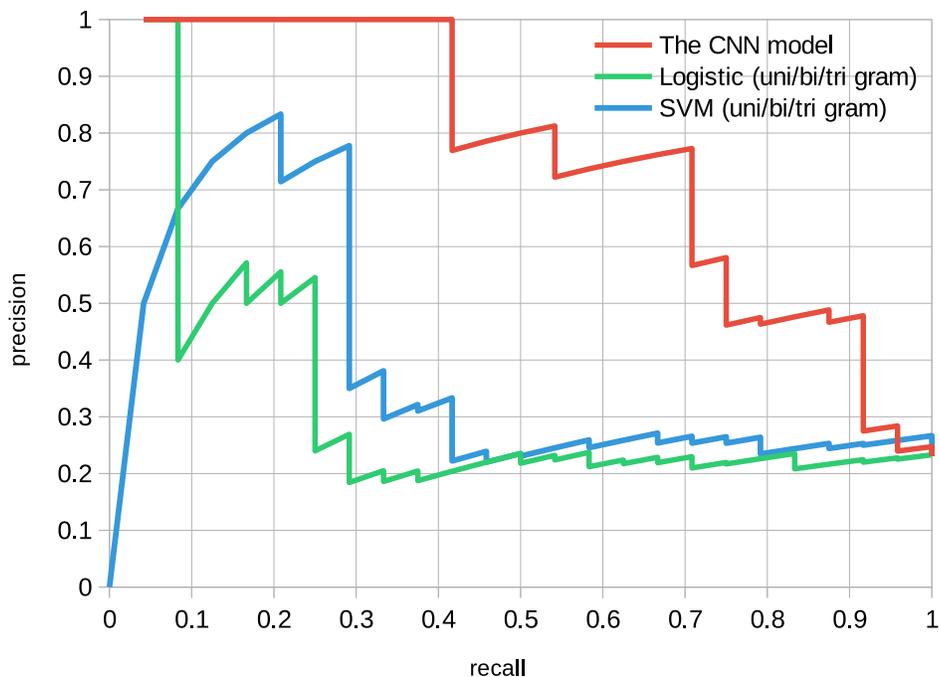}
	\caption{Precision-recall curve over the relationship data of the CNN model}
    \label{fig:precision_recall}
\end{figure}

Figure \ref{fig:precision_recall} shows the precision-recall curve of our CNN model. The precision was one when the recall was about 0.4, i.e., roughly speaking half the positive entity pairs were perfectly identified. The performance of the CNN model was superior to that of the baseline models. 

Table \ref{table:sample_snt} shows some representative sentences scored by the CNN model. A representative sentence is the highest scored sentence in a sentence set $S_{e_i, e_j}$ for each entity pair, i.e., a representative sentence is $s' = \textrm{argmax}_{s \in S_{e_i, e_j}} P( r = \textrm{True} | s )$, and it is the sentence that most likely describes the positive relation of the entity pair in the sentence set. The sentence and score indicate the grounds for the decision of the CNN model. The highly scored representative sentences seem to describe the desired relations (sentences 4 and 8) and, interestingly, relations described in the equation were also discovered by the model (sentences 2, 3, and 6). 
This implies that some important relations tend to be described in an equation.
This result also indicates that the relations in which we are interested are significantly different from typical relations in other NLP tasks like `has\_a', `is\_a'.

\begin{table}[th]
    \centering
	\caption{Sample representative sentences scored by the CNN model. Label P indicates that the entities are positively related in the test relationship data and label N indicates a negative relation. Entities in each sentence are underlined. The score is the $v_r z_2$ of each sentence.}
    \resizebox{\textwidth}{!}{\begin{tabular}{  c |c | p{43em} }
    \hline
    & Score/Label & Sentence \\ \hline
    1 & 36.5/P & ... the following \underline{matrix} \underline{form} : [11] $k \sim u = \lambda u$ ... \\ \hline
   2 & 34.8/P & ... $\delta c = r\sigma c / \tau$ is the characteristic or critical whisker length , $f$ and $r$ ... $\tau$ is the \underline{matrix} shear \underline{strength} ... \\ \hline
    3 & 34.2/P & ... \underline{toughness} ($\delta kcb$) and grain ... dvpwhere , $d$ is the \underline{matrix} ... \\ \hline
    4 & 31.0/P & ... \underline{cast} iron has a pearlite \underline{matrix} and ... \\ \hline
    5 & 28.6/P & after \underline{solution treatment}, the increase of grain size was not obvious because of the heat resistance introduced by ... .2 ) after aging ... .3 ) \underline{grain refining}, size reduction of ... \\ \hline
    6 & 26.0/N & \underline{solution strengthening} and precipitation strengthening respectively, ..., $\delta h-p$ was the yield \underline{strength} ... \\ \hline
    7 & 24.7/N & ...dislocation density in \underline{lath martensite} matrix due to the high content of element ... 100 steel delayed the recovery process during \underline{tempering} ... \\ \hline
    8 & 23.8/P & \underline{lath martensite} , which benefited the impact \underline{toughness} ... \\ 
    \vdots &  &   \\ 
    9 & -13.1/P &  ... the effect of ingot \underline{grain refinement} on the mechanical properties of al profiles which are manufactured through \underline{hot working} ... \\ \hline
    10 & -14.1/N & ... \underline{refining} the prior austenitic grain size ... \textsc{long context} ... the mechanical strength and \underline{cleavage resistance} ... \\ \hline
    11 & -16.4/N & ... enhanced solid \underline{solution strengthening} and composition \underline{homogenization} is larger than ... \\ \hline
    12 & -18.7/N & ... as the \underline{solution treatment} temperature increases to ..., the transformation ... and the formation of \underline{rim} o phase ... \\ \hline
    13 & -23.4/N & ... during the \underline{aging} treatment , the \underline{rim} o phase at the margin of $\alpha 2$ grains become ... \\ \hline
    \end{tabular}}
    \label{table:sample_snt}
\end{table}

\begin{table*}[ht]
	\caption{Source articles}
	\centering
    \resizebox{0.9\textwidth}{!}{
    
    \begin{tabular}{  c p{45em} } 
    1 &  Bin Wen and Nicholas Zabaras. Investigating variability of fatigue indicator parameters of two-phase nickel-based superalloy microstructures. DOI: \texttt{https://doi.org/10.1016/j.commatsci.2011.07.055} \\ 
    
    2 & Liguo Huang and Yuyong Chen. A study on the microstructures and mechanical properties of Ti–B20–0.1B alloys of direct rolling in the $\alpha$+$\beta$ phase region. DOI: \texttt{https://doi.org/10.1016/j.jallcom.2015.05.244} \\
    
    3 & Zengbin Yin, Juntang Yuan, Zhenhua Wang, Hanpeng Hu, Yu Cheng and Xiaoqiu Hu. Preparation and properties of an Al2O3/Ti(C,N) micro-nano-composite ceramic tool material by microwave sintering. DOI: \texttt{https://doi.org/10.1016/j.ceramint.2015.11.082} \\
    
	4 & Olamilekan Oloyede, Timothy D. Bigg, Robert F. Cochrane and Andrew M. Mullis. Microstructure evolution and mechanical properties of drop-tube processed, rapidly solidified grey cast iron. DOI: \texttt{https://doi.org/10.1016/j.msea.2015.12.020} \\
    
	5 & Chunchang Shi, Liang Zhang, Guohua Wu, Xiaolong Zhang, Antao Chen and Jiashen Tao. Effects of Sc addition on the microstructure and mechanical properties of cast Al-3Li-1.5Cu-0.15Zr alloy. DOI: \texttt{https://doi.org/10.1016/j.msea.2016.10.063} \\
    
	6 & Chenchong Wang, Chi Zhang, Zhigang Yang, Jie Su and Yuqing Weng. Microstructure analysis and yield strength simulation in high Co–Ni secondary hardening steel. DOI: \texttt{https://doi.org/10.1016/j.msea.2016.05.069} \\
    
	7 & Xiaohui Shi, Weidong Zeng, Qinyang Zhao, Wenwen Peng and Chao Kang. Study on the microstructure and mechanical properties of Aermet 100 steel at the tempering temperature around 482 $^\circ$C. DOI: \texttt{https://doi.org/10.1016/j.jallcom.2016.04.087} \\
    
	8 & H. Xie, L.-X. Du, J. Hu, G.-S. Sun, H.-Y. Wu and R.D.K. Misra. Effect of thermo-mechanical cycling on the microstructure and toughness in the weld CGHAZ of a novel high strength low carbon steel. DOI: \texttt{https://doi.org/10.1016/j.msea.2015.05.033} \\
    
	9 & Wei Haigen, Xia Fuzhong and Wang Mingpu. Effect of ingot grain refinement on the tensile properties of 2024 Al alloy sheets. DOI: \texttt{https://doi.org/10.1016/j.msea.2016.11.016} \\
    
	10 & A. Di Schino and C. Guarnaschelli. Effect of microstructure on cleavage resistance of high-strength quenched and tempered steels. DOI: \texttt{https://doi.org/10.1016/j.matlet.2009.06.032} \\
    
	11 & F.L. Cheng, T.J. Chen, Y.S. Qi, S.Q. Zhang and P. Yao. Effects of solution treatment on microstructure and mechanical properties of thixoformed Mg2Sip/AM60B composite. DOI: \texttt{https://doi.org/10.1016/j.jallcom.2015.02.147} \\
    
	12, 13 &  X. Chen, F.Q. Xie, T.J. Ma, W.Y. Li and X.Q. Wu. Microstructural evolution and mechanical properties of linear friction welded Ti2AlNb joint during solution and aging treatment. DOI: \texttt{https://doi.org/10.1016/j.msea.2016.05.030} \\

    \end{tabular}
    }
\end{table*}

\section{End-to-end system} \label{sec:end_to_end}
\begin{figure}[t]
  \begin{tabular}{cc}
  \begin{minipage}{0.40\hsize}
     \begin{center}
        \includegraphics[width=\textwidth]{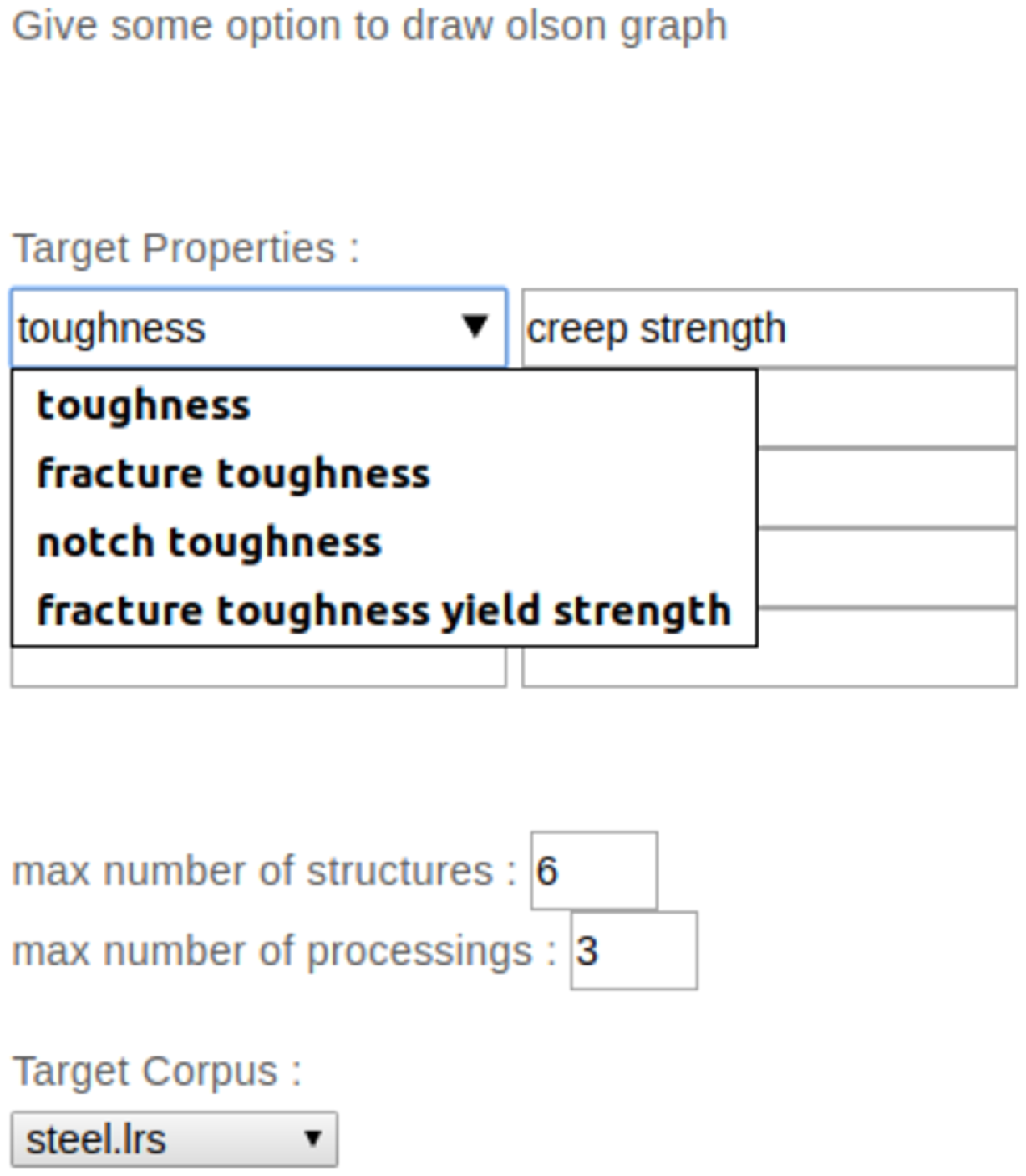}
     \end{center}
  \end{minipage} &
  \begin{minipage}{0.50\hsize}
     \begin{center}
        \includegraphics[width=\textwidth]{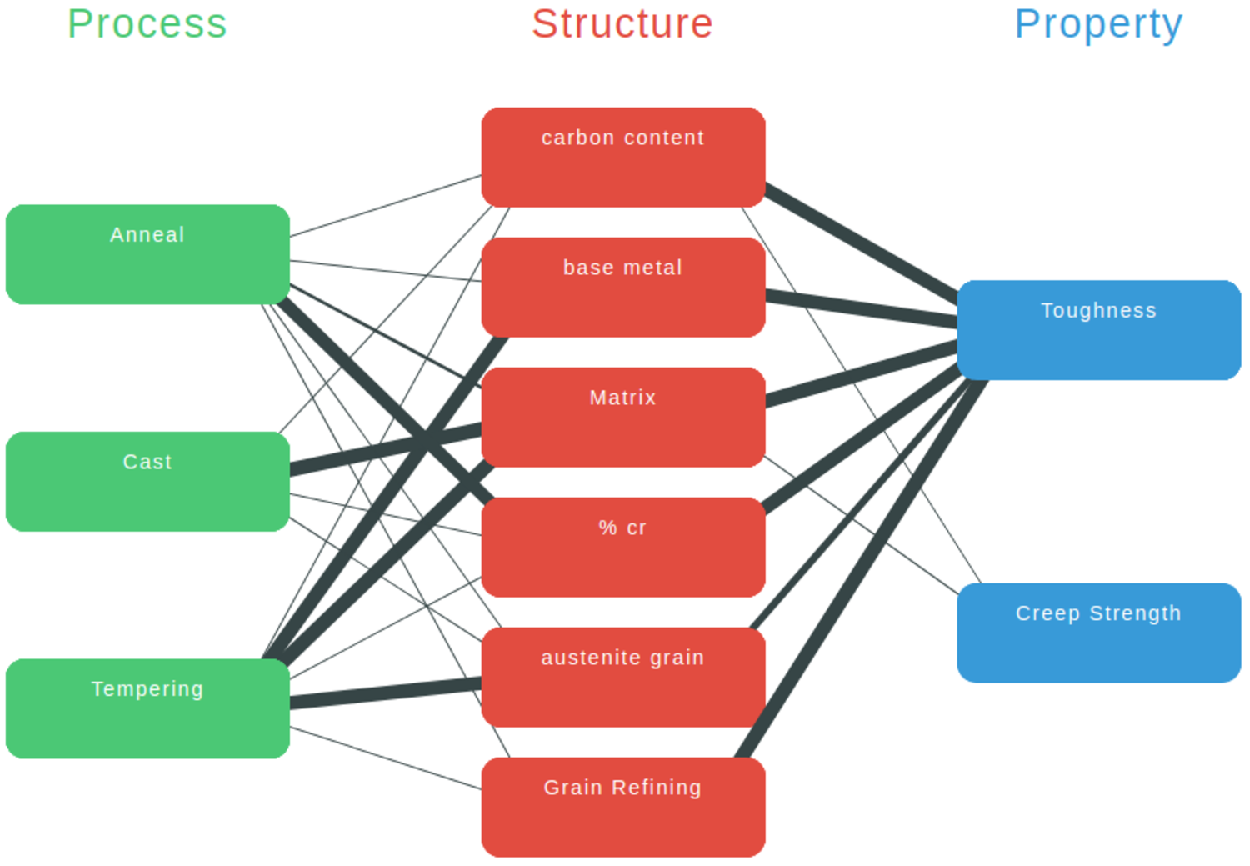}
     \end{center}
  \end{minipage} \\ \\
  \end{tabular}
  \begin{center}
  \begin{minipage}{0.90\hsize}
        \includegraphics[width=\textwidth]{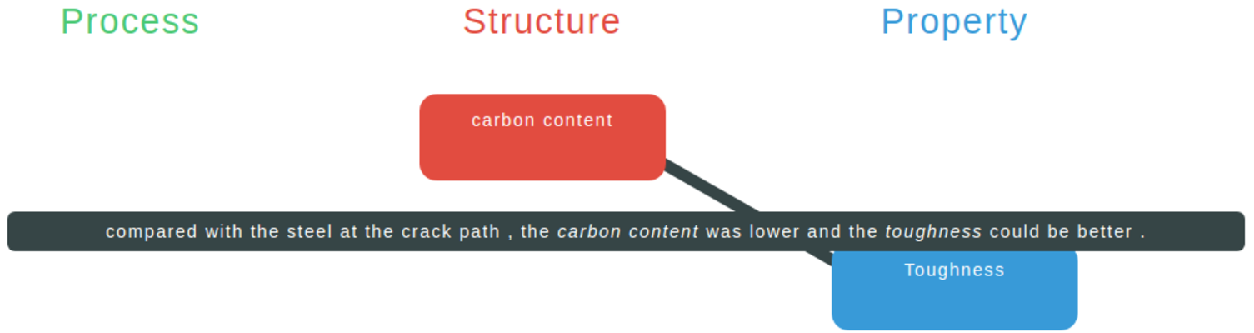}
  \end{minipage} 
  \end{center}
  \caption{The end-to-end demo system. a) Desired properties and a base material were selected. b) A sample of the generated PSPP design chart. The desired properties were \textit{toughness} and \textit{creep strength}, and `steel' was selected as base material. c) A sentence describing the relation between \textit{toughness} and \textit{carbon content}.}
  \label{fig:demo}
\end{figure}

We developed a web-based end-to-end demo system to demonstrate our system in Fig.~\ref{fig:demo}. The demo system worked in a typical scenario of material development, where a scientist was looking for factors related to certain desired properties. The demo system provides an PSPP design chart for the properties that the scientist provided. The end-to-end system works on Apache Tomcat\footnote{http://tomcat.apache.org}.

The system input consisted of the desired properties along with a base material. The desired properties were selected from a list of properties collected as in Section~\ref{sec:factor}. The base material was the target material, such as aluminum or titanium. It was important to obtain the desired knowledge. For example, the relationship between `strength' and `matrix' in titanium alloys might have been different from this relationship in aluminum alloys. 

Then, the system predicts PSPP relations from the scientific articles about the base material. Firstly, the system selects a set of scientific articles for the base material. As in Section~\ref{sec:corpus}, the articles were collected by keyword search in ScientceDirect. Thus the system predicted all relations among the entities collected in Section~\ref{sec:factor}, and scored them as in Section~\ref{sec:model}. Then the system generated a PSPP chart for the given properties as Section~\ref{sec:graph_completion}.

The system output was a PSPP design chart suggesting the required structures and processes. The chart formed by three columns --process, structure, and property-- suggested relations from the processes to the desired properties. Moreover, for each relation, the system provided a representative sentence to justify the relation and aid the researcher's understanding.

\section{Conclusions and contribution} \label{sec:conclusion}

In this study, we developed and tested our knowledge extraction and representation system intended to support material design, by representing knowledge as relationships. Knowledge was represented as relationships in PSPP design charts. We leveraged weakly supervised learning for relation extraction. The end-to-end system proved our concept, and its relation extraction performance was superior to that of other baseline models.

Our contribution in this study is twofold.
Firstly, we proposed a novel knowledge graph based on PSPP charts, and developed a system to build the knowledge graph from text using NLP technologies.
Secondly, we experimentally verified that such technical knowledge can be extracted from text by using machine learning models. Our target knowledge is relations in PSPP design charts. These relations appear rather technical and significantly different from typical relations in NLP such as `has\_a' and `is\_a'. Extraction of these relations from text is nontrivial, and might need other knowledge resource such as equations and properties of materials; however, we experimentally verified that a state-of-the-art machine learning model can find these relations from texts.

\section{Follow-up work}
Knowledge graphs in the scientific domain are recently highly demanded, and numerous works have been published. We overview the related work after our work.  

As we employed the PSPP reciprocity, various types of knowledge graphs are studied for each target information. In the general scientific domain, \citet{10.1145/3227609.3227689} proposed the vision and infrastructure of a knowledge base for the general scientific domain. In biology, \citet{10.1145/3292500.3330942} pointed out some relational scientific facts are true under specific conditions in biology. For example, given the following sentence:
\begin{centering}
	``We observed that ... alkaline pH increases the activity of TRV5/V6 channels in Jurkat T Cells.''~\citep{PMID:26096460}
\end{centering}
We can find a relational fact, \{``alkaline pH'', \textit{increase}, ``TRV5/V6 channels''\}, which is true if \{``TRV5/V6 channels'', \textit{locate}, ``Jurkat T Cells'' \}. Another knowledge base for biology combines multiple structured databases and scientific papers~\citep{Manica2019AnIE}. In materials science, \citet{Mrdjenovich2020} manually developed Propnet consisting of 115 material properties and 69 relationships, and \citet{strotgen2019towards} proposed the Bosche Materials Science Knowledge Base consisting of 40K relational facts for solid oxide fuel cells. 

Unlike we find mentions, tokens referring an entity, by using heuristic string matching, recently mention-level annotations are available in some tasks for the general scientific domain. 
For example, SemEval 2017 ScientificIE~\citep{semeval2017task10} and SemEval-2018: ``Semantic Relation Extraction and Classification in Scientific Papers''~\citep{gabor-etal-2018-semeval} consists of three tasks; a) mention identification b) mention classification c) mention-level relation extraction, and each mention, the class of each mention, and their relations are labeled in the training data. Additionally, \citet{luan-etal-2018-multi} proposed SciERC as extending these datasets. These annotations provide cleaner training labels and make training efficient. 

Information extraction for materials science (material informatics) is also highly demanded and actively studied. 
For example, another desired information to be extracted for materials science might be synthesis procedures. A synthesis procedure is a sequence of operations to synthesis a compound. Mention-level annotated datasets are provided for this task~\citep{friedrich-etal-2020-sofc, mysore-etal-2019-materials, Kononova2019}, and \citet{DBLP:journals/corr/abs-1711-06872} apply the generative model of \citet{kiddon-etal-2015-mise} to induce the procedures. 
Furthermore, several essential NLP technologies are studied for material informatics, such as entity recognition for materials science~\citep{Weston2019, kim2017}, and word2vec~\citep{DBLP:journals/corr/abs-1301-3781} on materials science publications~\citep{tshitoyan2019}.

\chapter{Conclusion} \label{sec:thesis_conclusion}
In this thesis, we discussed reading comprehension, focusing on entities and their relations. We started with an overview of reading comprehension tasks and the role of entities and their relations in these tasks. In early work, these tasks provide a small hand-written dataset for rule-based systems. Later, the datasets are getting bigger and bigger for machine learning models, especially for deep neural network models that are capable of being trained on such large scale training data. Then we claim that the goal of these tasks is to test the reading comprehension skills of machines, and it differentiates the reading comprehension from other question answering tasks. Additionally, we are interested in not only testing these skills but also how the machine understands texts, and then claim that entities and their relation can be a key to explain it.

In Chapter \ref{sec:work1}, we constructed a reading comprehension dataset, WDW, that is designed to validate the reading comprehension skills, especially the skill to understand entities in given texts. Here we used baseline systems and a sampling approach to control the difficulty of the dataset so that each question requires appropriate reading comprehension skills to solve it. The dataset gives a larger gap between human performance and machine performance, which shows that our dataset requires deeper text understanding. 

In Chapter \ref{sec:work2}, we investigated the skill to understand entities and experimentally identified a neural network module that associates with each entity in neural readers. We explored neural readers and classified them into aggregation readers and explicit readers by their neural structures on top of contextual token embeddings. We experimentally found contextual token embeddings that strongly correlate with each entity, and then showed the attention layer of the aggregation reader mimics the explicit reference of the explicit reader.

In Chapter \ref{sec:wikihop}, we feedbacked the findings to another entity and relation centric reading comprehension dataset, Wikihop, and improved the performance of the neural network model. Here we leverage the neural structure associating with each entity for scoring each candidate answer. Additionally, we proposed a training algorithm that can train self-attention layers without quadratically consuming the memory. 

In Chapter \ref{sec:work3}, we developed a visualization system that summarizes given texts into a graph consisting of entities and their relations. This system extracts entities and their relations from a bunch of scientific articles. These entities and relations produce a graph that visualizes a summary of the given scientific articles. This work is collaborative work with materials science, and our target information to be visualized is PSPP relations. We showed that such highly scientific relations could be extracted by the novel neural network trained on about 100 labeled relations and scientific articles.

\section{Future work}

We presented our contribution to reading comprehension focusing on entities and their relations. Here, we discuss straightforwardly more work to do to understand the reading comprehension skills of deep neural networks better.

Thanks to the deep neural networks and large scale datasets, the performance of machines in reading comprehension tasks is significantly improved. On the other hand, it becomes more and more difficult to explain each semantic role of vector representation as the network structure becomes more and more complicated.

We presented an empirical analysis of neural readers in Chapter \ref{sec:work2}, and identified contextual token embeddings that strongly correlate with each entity embedding in an entity-centric dataset. A follow-up question might be the following. 
\begin{center}
``How are entities treated in other reading comprehension styles and other neural models ?''
\end{center}
Recently, other reading comprehension styles, such as the span prediction and free-form answer, is more popular, and other neural models are proposed, such as Transformer. However, they are still based on linear transformations; thus, we can capture a correlation between arbitrary two vector representations by computing inner-product just as Chapter \ref{sec:work2}.  Then, we can apply the same approach to these reading comprehension styles and capture neural module that correlates with each entity. 

We are also interested in a practical issue of the machine learning we faced in Chapter \ref{sec:work3}, a lacking of training data for a specific domain. In many practical cases, it is untrivial to collect enough amount of manually labeled training data for neural network models, and a domain-specific dataset tends to be smaller than a general-domain dataset, like~\citep{deng2009imagenet}. Thus, the size of the dataset tends to be a bottleneck of the performance. In this thesis, we took three approaches to address this issue. Firstly, we build a dataset by heuristically matching news articles and sampling them in Chapter \ref{sec:work1}. Secondly, we initialize our model with a pre-trained neural network and then fine-tuned in Chapter \ref{sec:wikihop}. Thirdly, we combined relational information and texts by the idea of distant supervision in Chapter \ref{sec:work3}. There are other interesting approaches, including zero-shot learning~\citep{5206594}, one-shot learning~\citep{1597116, NIPS2004_2576} few-shot learning~\citep{yu-etal-2018-diverse}. We believe it is critical to choose a suitable learning scheme to develop a domain-specific machine learning system.

%
%
\bibliographystyle{mynat}
\bibliography{material,mywork,main}

\end{document}